\documentclass[runningheads]{llncs}

\usepackage{graphicx}
\usepackage{amsmath,amssymb} %
\usepackage{color}
\usepackage{booktabs} %
\usepackage{multirow}
\usepackage{xspace}
\usepackage{bm}
\usepackage{enumitem}
\usepackage{url}

\newcommand{\etal}{\textit{et al}.}
\newcommand{\funit}{FUNIT\xspace}
\newcommand{\cocofunit}{COCO-FUNIT\xspace}
\newcommand{\coco}{COCO\xspace}

\usepackage{subcaption}
\newcommand{\mysubsec}[1]{\medskip\noindent{\bf #1}}
\newcommand{\mysec}[1]{\section{#1}}

\usepackage[width=122mm,left=12mm,paperwidth=146mm,height=193mm,top=12mm,paperheight=217mm]{geometry}

\begin{document}

\pagestyle{headings}
\mainmatter
\def\ECCV18SubNumber{467} 

\title{COCO-FUNIT:  \\ Few-Shot Unsupervised Image Translation with \\
a Content Conditioned Style Encoder}

\titlerunning{COCO-FUNIT}

\authorrunning{Saito et al.}

\author{Kuniaki Saito$^{1,2}$ \and
Kate Saenko$^{1}$\and
Ming-Yu Liu$^{2}$}
\authorrunning{Saito et al.}
\institute{Boston University$^{1}$ \hspace{0.5in} NVIDIA$^{2}$ \\
\email{\{keisaito,saenko\}@bu.edu}, \ \email{mingyul@nvidia.com}}
\maketitle

\begin{figure}
    \centering
    \includegraphics[width=\textwidth]{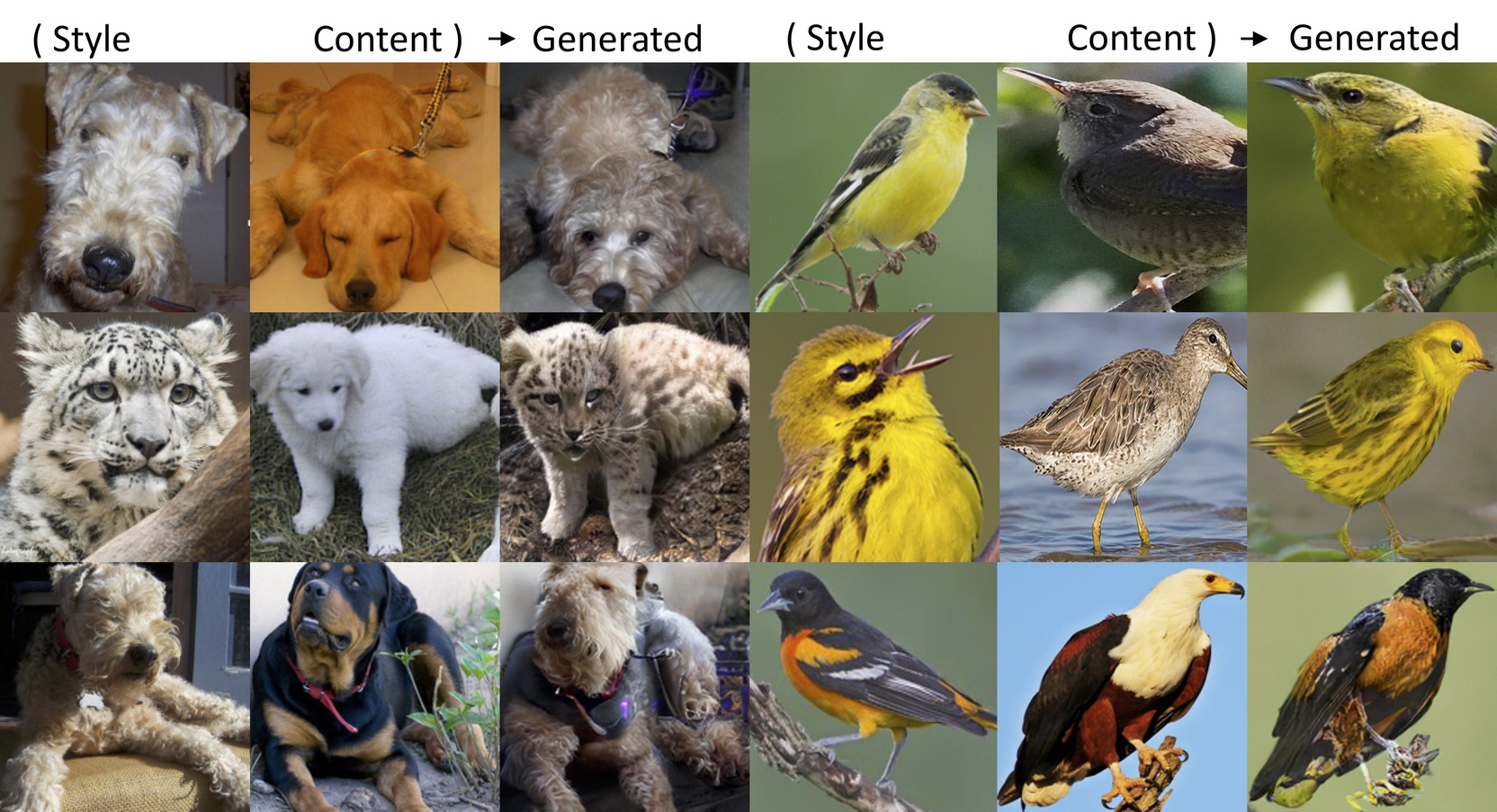}
    \caption{
    Given as few as one style example image from an object class unseen during training, our model can generate a photorealistic translation of the input content image in the unseen domain.}
    \label{fig:teaser}
\end{figure}

\begin{abstract}
	Unsupervised image-to-image translation intends to learn a mapping of an image in a given domain to an analogous image in a different domain, without explicit supervision of the mapping. Few-shot unsupervised image-to-image translation further attempts to generalize the model to an unseen domain by leveraging example images of the unseen domain provided at inference time. While remarkably successful, existing few-shot image-to-image translation models find it difficult to preserve the structure of the input image while emulating the appearance of the unseen domain, which we refer to as the \textit{content loss} problem. This is particularly severe when the poses of the objects in the input and example images are very different. To address the issue, we propose a new few-shot image translation model, \cocofunit, which computes the style embedding of the example images conditioned on the input image and a new module called the constant style bias. Through extensive experimental validations with comparison to the state-of-the-art, our model shows effectiveness in addressing the \textit{content loss} problem. For code and pretrained models, please check out
	\url{https://nvlabs.github.io/COCO-FUNIT/}.
\end{abstract}

\mysec{Introduction}\label{sec:intro}

\begin{figure*}[tbh!]
    \centering
    \includegraphics[width=\linewidth]{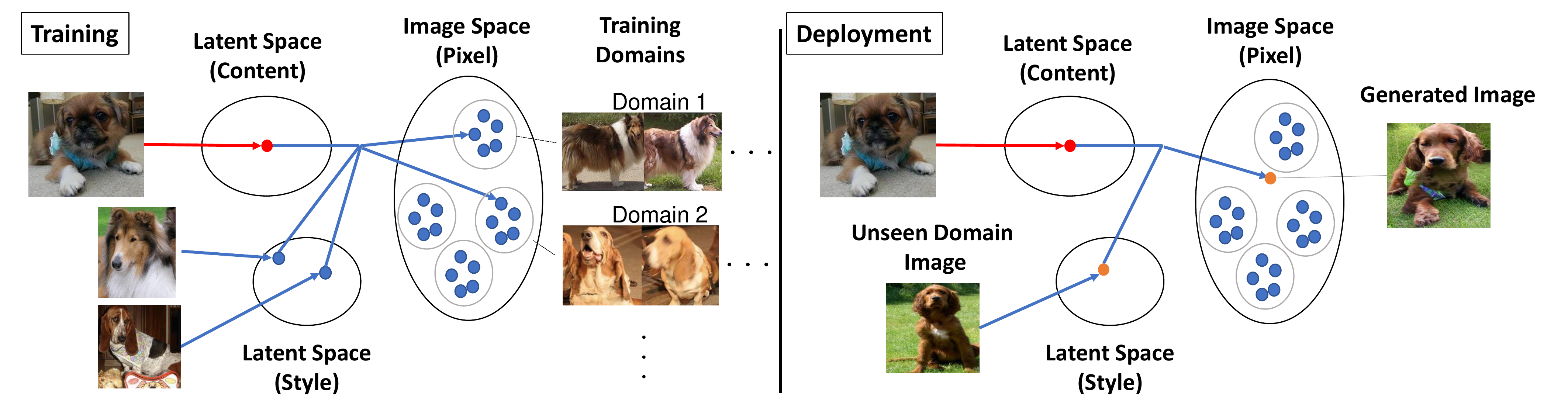}
    \caption{Few-shot image-to-image translation. \textbf{Training.} The training set consists of many domains. We train a model to translate images between these domains. \textbf{Deployment.} We apply the trained model to perform few-shot image translation. Given a few examples from a test domain, we aim to translate a content image into an image analogous to the test class.}    
    \label{fig:setting}
\end{figure*}

Image-to-Image translation~\cite{isola2017image,wang2018high} concerns learning a mapping that can translate an input image in one domain into an analogous image in a different domain. Unsupervised image-to-image translation~\cite{zhu2017unpaired,liu2017unsupervised,liang2017dual,kim2017learning,liu2016coupled,taigman2017unsupervised,choi2017stargan} attempts to learn such a mapping without paired data. Thanks to the introduction of novel network architectures and learning objective terms, the state-of-the-art has advanced significantly in the past few years. However, while existing unsupervised image-to-image translation models can generate realistic translations, they still have several drawbacks. First, they require a large amount of images from the source and target domains for training. Second, they cannot be used to generate images in unseen domains. These limitations are addressed in the few-shot \textit{unsupervised} image-to-image translation framework~\cite{liu2019few}. By leveraging example-guided episodic training, the few-shot image translation framework~\cite{liu2019few} learns to extract the domain-specific style information from a few example images in the unseen domain during test time, mixes it with the domain-invariant content information extracted from the input image, and generates a few-shot translation output as illustrated in Fig.~\ref{fig:setting}.

\begin{figure}[tbh!]
    \centering
    \includegraphics[width=\textwidth]{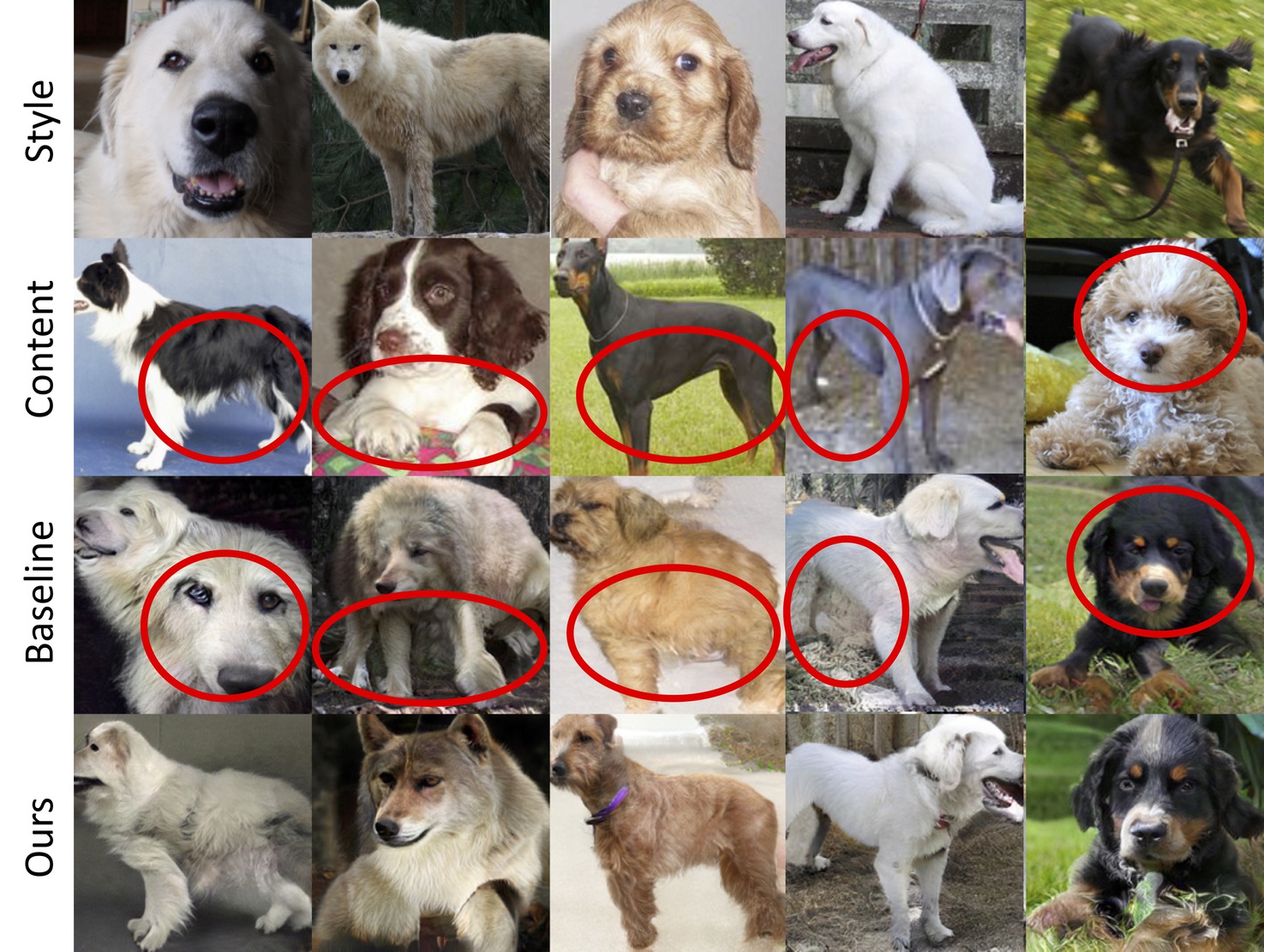}

    \caption{Illustration of the \textit{content loss} problem. The images generated by the baseline~\cite{liu2019few} fail to preserve domain invariant appearance information in the content image. The animals' bodies are sometimes merged with the background (column 3, \& 4), scales of the generated body parts are sometimes inconsistent with the input (column 5), and some body parts absent in the content image show up (column 1 \& 2). Our proposed method solves this ``content loss'' problem.}
    \label{fig:content_loss}

\end{figure}

However, despite showing encouraging results on relatively simple tasks such as animal face and flower translation, the few-shot translation framework~\cite{liu2019few} frequently generates unsatisfactory translation outputs when the model is applied to objects with diverse appearance, such as animals with very different body poses. Often, the translation output is not well-aligned with the input image. The domain invariant content that is supposed to remain unchanged disappears after translation, as shown in Fig.~\ref{fig:content_loss}. We will call this issue the \textit{content loss} problem. We hypothesize that solving the content loss problem would produce more faithful and photorealistic few-shot image translation results.

But why does the content loss problem occur? To learn the translation in an unsupervised manner, Liu~\etal~\cite{liu2019few} rely on inductive bias injected by the network design and adversarial training~\cite{goodfellow2014generative} to transfer the appearance from the example images in the unseen domain to the input image. However, as there is no supervision, it is difficult to control what to be transferred precisely. Ideally, the transferred appearance should contain just the style. In reality, it often contains other information, such as the object pose. 

In this paper, we propose a novel network architecture to counter the content loss problem. We design a style encoder called the \textit{content-conditioned style encoder}, to hinder the transmission of task-irrelevant appearance information to the image translation process. In contrast to the existing style encoders, our style code is computed by conditioning on the input content image. We use a new architecture design to limit the variance of the style code. We conduct an extensive experimental validation with a comparison to the state-of-the-art method using several newly collected and challenging few-shot image translation datasets. Experimental results, including both automatic performance metrics and user studies, verify the effectiveness of the proposed method in dealing with the content loss problem.

\mysec{Related Works}

{\bf Image-to-image translation}. Most of the existing models are based on the Generative Adversarial Network (GAN)~\cite{goodfellow2014generative} framework. Unlike unconditional GANs~\cite{goodfellow2014generative,karras2019style,karras2017progressive,gulrajani2017improved,mao2017least}, which learn to map random vectors to images, existing image-to-image translation models are mostly based on conditional GANs where they learn to generate a corresponding image in the target domain conditioned on the input image in the source domain. Depending on the availability of paired input and output images as supervision in the training dataset, image-to-image translation models can be divided into supervised~\cite{isola2017image,wang2018high,park2019semantic,liu2019learning,chen2017photographic,qi2018semi,zhu2017toward,zhu2017your,zhao2019image,wang2018discriminative,wang2018video,wang2019example} or unsupervised~\cite{zhu2017unpaired,liu2017unsupervised,liang2017dual,kim2017learning,liu2016coupled,taigman2017unsupervised,choi2017stargan,pumarola2018ganimation,huang2018multimodal,lee2018diverse,shen2019towards,gokaslan2018improving,benaim2018one,albahar2019guided}. Our work falls in the category of unsupervised image-to-image translation. However, instead of learning a mapping between two specific domains, we aim at learning a flexible mapping that can be used to generate images in many unseen domains. Specifically, the  mapping is only determined at test time, via example images. When using example images from a different unseen domain, the same model can generate images in the new unseen domain.

{\bf Multi-domain image translation}. Several works~\cite{choi2017stargan,anoosheh2017combogan,hui2017unsupervised,choi2019stargan} extend the unsupervised image translation to multiple domains. They learn a mapping between multiple seen domains, simultaneously. Our work differs from the multi-domain image translation works in that we aim to translate images to \textit{unseen} domains.

\textbf{Few-shot image translation.} Several few-shot methods are proposed to generate human images~\cite{wang2019few,wang2019example,Siarohin2019monkeynet,han2018viton}, scenes~\cite{wang2019example}, or human faces~\cite{zakharov2019few,wang2019few,gu2019ladn} given a few instances and semantic layouts in a test time. These methods operate in the supervised setting. During training, they assume access to paired input (layout) and output data. Our work is most akin to the \funit work~\cite{liu2019few} as we aim to learn to generalize the translation to unseen domain without paired input and output data. We build on top of the \funit work where we first identify the \textit{content loss} problem and then address it with a novel content-conditioned style encoder architecture.

{\bf Example-guided image translation} refers to methods that generate a translation of an input conditioning on some example images. Existing works in this space~\cite{huang2018multimodal,park2019semantic,liu2019few} use a style encoder to extract style information from the example images. Our work is also an example-guided image translation method. However, unlike the prior works where the style code is computed independent of the input image, our style code is computed by conditioning on the input image, where we normalize the style code using the content to 
prevent over-transmission of the style information to the output.

{\bf Neural style transfer} studies approaches to transfer textures from a painting to a real photo. While existing neural style transfer methods~\cite{gatys2015texture,huang2017adain,li2017universal} can generalize to unseen textures, they cannot generalize to unseen shapes, necessary for image-to-image translation. Our work is inspired by these works, but we focus on generalizing the generation of both unseen shapes and textures, which is essential to few-shot unsupervised image-to-image translation.

\mysec{Method}

In this section, we start with a brief explanation of the problem setup, introduce the basic architecture, and then describe our proposed architecture. Throughout the paper, the two words, "class" and "domain", are used interchangeably since we treat each object class as a domain.

\mysubsec{Problem setting.} Fig.~\ref{fig:setting} provides an overview of the few-shot image translation problem~\cite{liu2019few}. Let $X$ be a training set consists of images from $K$ different domains. For each image in $X$, the class label is known. Note that we operate in the unsupervised setting where corresponding images between domains are \textit{unavailable}. The few-shot image-to-image translation model learns to map a ``content'' image in one domain to an analogous image in the domain of the input ``style'' examples. In the test phase, the model sees a few example images from an unseen domain and performs the translation. 

\begin{figure*}[!t]
    \centering
    \includegraphics[width=0.99\linewidth]{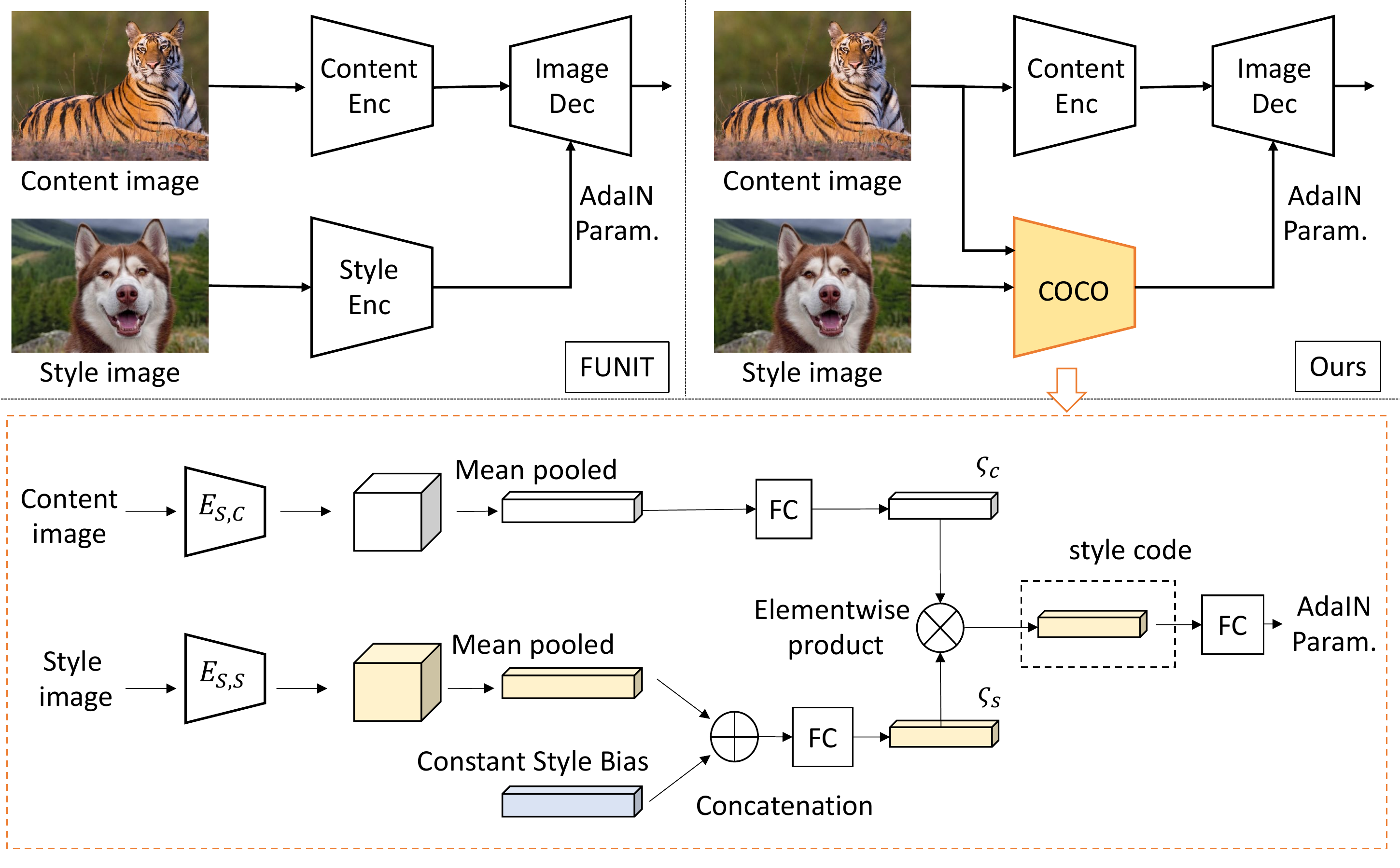}
    \caption{\textbf{Top}. The \funit baseline~\cite{liu2019few} vs. our \cocofunit. To highlight, we use a novel style encoder called the content-conditioned style encoder where the content image is also used in computing the style code for few-shot unsupervised image-to-image translation. \textbf{Bottom}. Detail design of the content-conditioned style encoder. Please refer to the main text for more details.}
    \label{fig:model}
\end{figure*}

During training, a pair of content and style images ${x_c, x_k}$ is randomly sampled. Let $x_k$ denote a style image in domain $k$. The content image $x_c$ can be from any domains in $K$. The generator $G$ translates $x_c$ into an image of class $k$ ($\bar{\bm{x}}_{k}$) while preserving the content information of $x_c$.
\begin{equation}\label{eq:base}
    \bar{\bm{x}}_{k} = G(x_c, x_{k})
\end{equation}

In the test phase, the generator takes style images from a domain unseen during training, which we call the target domain. The target domain can be any related domain, not included in $K$. 

\mysubsec{\funit baseline.} \funit uses an example-guided conditional generator architecture as illustrated in the top-left of Fig.~\ref{fig:model}. It consists of three modules, 1) content encoder $E_c$, 2) style encoder $E_s$, and 3) image decoder $F$. $E_c$ takes content image $x_c$ as input and outputs content embedding $z_c$. $E_s$ takes style image $x_s$ as input and output style embedding $z_s$. Then, $F$ generates an image using $z_c$ and $z_s$, where $z_s$ is used to generate the mean and scale parameters of adaptive instance normalization (AdaIN) layers \cite{huang2017adain} in $F$. The AdaIN design is based on the assumption that the domain-specific information can be governed by the first and second order statistics of the activation and has been used in several GAN frameworks~\cite{huang2018multimodal,liu2019few,karras2019style}. We further note that when multiple example/style images are present. \funit extracts a style code from each image and uses the average style code as the final input to $F$. To sum up, in \funit the image translation is formalized as follows, 

\begin{equation}\label{eq:funit}
z_c = E_c(x_c),\ \ z_s = E_s(x_s),\ \ \bar{\bm{x}} = F(z_c, z_s).
\end{equation}

\begin{figure}[tbh!]
    \centering
    \includegraphics[width=0.99\textwidth]{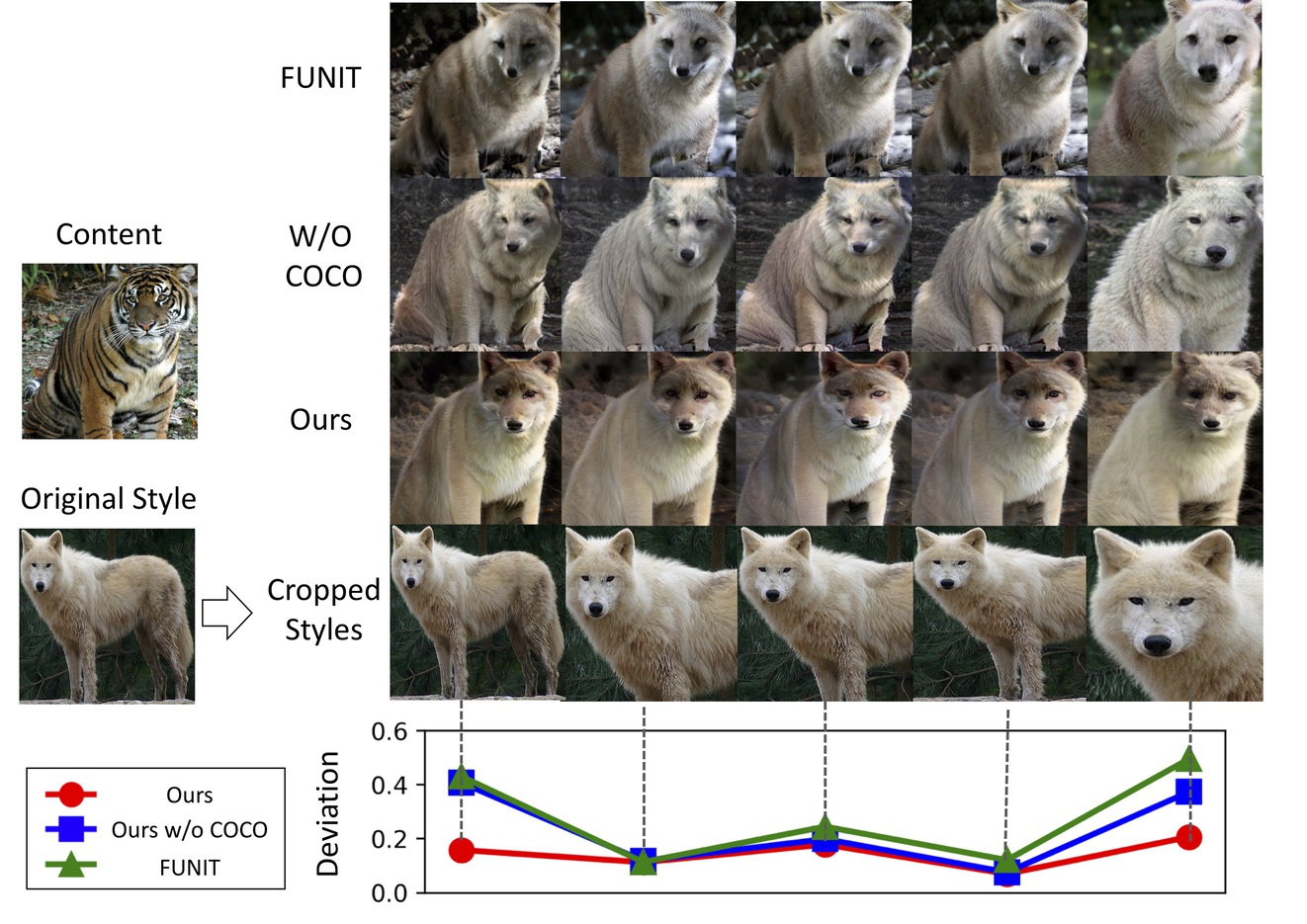}
    \caption{We compare variations of the computed style codes due to variations in the style images for different methods. Note that for a fair comparison, in addition to the original \funit baseline~\cite{liu2019few}, we create an improved \funit method by using our improved design for the content encoder, image decoder, and discriminator, which is termed "Ours w/o COCO". "Ours" is our full algorithm where we use COCO as a drop-in replacement for the style encoder in the \funit framework. In the bottom part of the figure, we plot the variations of the style code due to using different crops of a style image. Specifically, the style code for each style image is first extracted for each method. We then compute the mean of the style codes for each method. The magnitudes of the deviations from the mean style code are then plotted. Note that to calibrate the network weights in different methods, all the style codes are first normalized by the mean extracted from 500 style images for each method. As shown in the figure, "Ours" produces more consistent translation outputs, which is a direct consequence of a more consistent style code extraction mechanism.}
    \label{fig:style_variation_analysis}
\end{figure}

\mysubsec{Content loss.} As illustrated in Fig.~\ref{fig:content_loss}, the \funit method suffers from the content loss problem---the translation result is not well-aligned with the input image. While a direct theoretical analysis is likely elusive, we conduct an empirical study, aiming at identify the cause of the content loss problem. As shown in Fig.~\ref{fig:style_variation_analysis}, we compute different translation results of a content image based on a different style image where each of the style images is cropped from the same original style image. In the plot, we show variations of the deviation of the extracted style code due to different crops. Ideally, the plot should be constant as long as the crop covers sufficient appearance signature of the target class since that should be all required to generate a translation in the unseen domain. However, the \funit style encoder produces very different style codes as using different crops. Clearly, the style code contains other information about the style image such as the object pose. We hypothesize this is the cause of the content loss problem and revisit the translator network design for addressing it.

\mysubsec{Content-conditioned style encoder (COCO).} We hypothesize that the content loss problem can be mitigated if the style embedding is more robust to small variations in the style image. To this end, we design a new style encoder architecture, called the COntent-COnditioned style encoder (\coco). There are several distinctive features in \coco. The most obvious one is the conditioning in the content image as illustrated in the top-right of Fig.~\ref{fig:model}. Unlike the style encoder in \funit, \coco takes \textit{both} content and style image as input. With this content-conditioning scheme, we create a \textit{direct} feedback path during learning to let the content image influence how the style code is computed. It also helps reduce the direct influence of the style image to the extract style code.

The bottom part of Fig.~\ref{fig:model} details the \coco architecture. First, the content image is fed into encoder $E_{S,C}$ to compute a spatial feature map. This content feature map is then mean-pooled and mapped to a vector $\zeta_c$. Similarly, the style image is fed into encoder $E_{S,S}$ to compute a spatial feature map. The style feature map is then mean-pooled and concatenated with an input-independent bias vector, which we refer to as the constant style bias (CSB). Note that while the regular bias in deep networks is added to the activations, in CSB, the bias is concatenated with the activations. The CSB provides a fixed input to the style encoder, which helps compute a style code that is less sensitive to the variations in the style image. In the experiment section, we show that the CSB can also be used to control the type of appearance information that is transmitted from the style image. When the CSB is activated, mostly texture-based appearance information is transferred. Note that the dimension of the CSB is set to 1024 through the paper.

The concatenation of the style vector and the CSB is mapped to a vector $\zeta_s$ via a fully connected layer. We then perform an element-wise product operation to $\zeta_c$ and $\zeta_s$, which is our final style code. The style code is then mapped to produce the AdaIN parameters for generating the translation. Through this element-wise product operation, the resulting style code is heavily influenced by the content image. One way to look at this mechanism is that it produces a customized style code for the input content image.

We use the COCO as a drop-in replacement for the style encoder in \funit. Let $\phi$ denote the COCO mapping. The translation output is then computed via
\begin{equation}\label{eq:encode}
z_c = E_c(x_c),\ z_s = \phi(E_{s,s}(x_s), E_{s,c}(x_c)),\ \bar{\bm{x}} = F(z_c, z_s).
\end{equation}
As shown in Fig.~\ref{fig:style_variation_analysis}, the style code extracted by the COCO is more robust to variations in the style image. Note that we set $E_{S,C}\equiv E_C$ to keep the number of parameters in our model similar to that in \funit.

We note that the proposed \coco architecture shows only one way to generate the style code conditioned on the content and to utilize the CSD. Certainly, there exist other design choices that could potentially lead to better translation performance. However, since this is the first time these two components are used for the few-shot image-to-image translation task, we focus on analyzing their contribution in one specific design, i.e., our design. An exhaustive exploration is beyond the scope of the paper and is left for future work.

In addition to the \coco, we also improve the design of the content encoder, image decoder, and discriminator in the \funit work~\cite{liu2019few}. For the content encoder and image decoder, we find that replacing the vanilla convolutional layers in the original design with residual blocks~\cite{he2016deep} improves the performance so does replacing the multi-task adversarial discriminator with the project-based discriminator~\cite{miyato2018cgans}. In Appendix~\ref{sec::additional_results}, we report their individual contribution to the few-shot image translation performance.

\mysubsec{Learning.} We train our model using three objective terms. We use the GAN loss ($\mathcal{L}_{\text{GAN}}(D,G)$) to ensure the realism of the generated images given the class of the style images. We use the image reconstruction loss ($\mathcal{L}_{\text{R}}(G)$) to encourage the model to reconstruct images when both the content and the style are from the same domain. We use the discriminator feature matching loss ($\mathcal{L}_{\text{FM}}(G)$) to minimize the feature distance between real and fake samples in the discriminator feature space, which has the effect of stabilizing the adversarial training and contributes to generating better translation outputs as shown in the \funit work. In Appendix~\ref{sec::learning}, we detail the computation of each loss. 
Overall the objective is 
\begin{align}
\min_{D}\max_{G} \mathcal{L}_{\text{GAN}}(D,G) + 
\lambda_{\text{R}} \mathcal{L}_{\text{R}}(G) + \lambda_{\text{F}} \mathcal{L}_{\text{FM}}(G), 
\end{align}
where $\lambda_{\text{R}}$ and $\lambda_{\text{F}}$ denote trade-off parameters for two losses. We set $\lambda_{\text{R}}$ 0.1 and $\lambda_{\text{F}}$ 1.0 in all of the experiments.

\begin{figure}[!t]
    \centering
    \includegraphics[width=\textwidth]{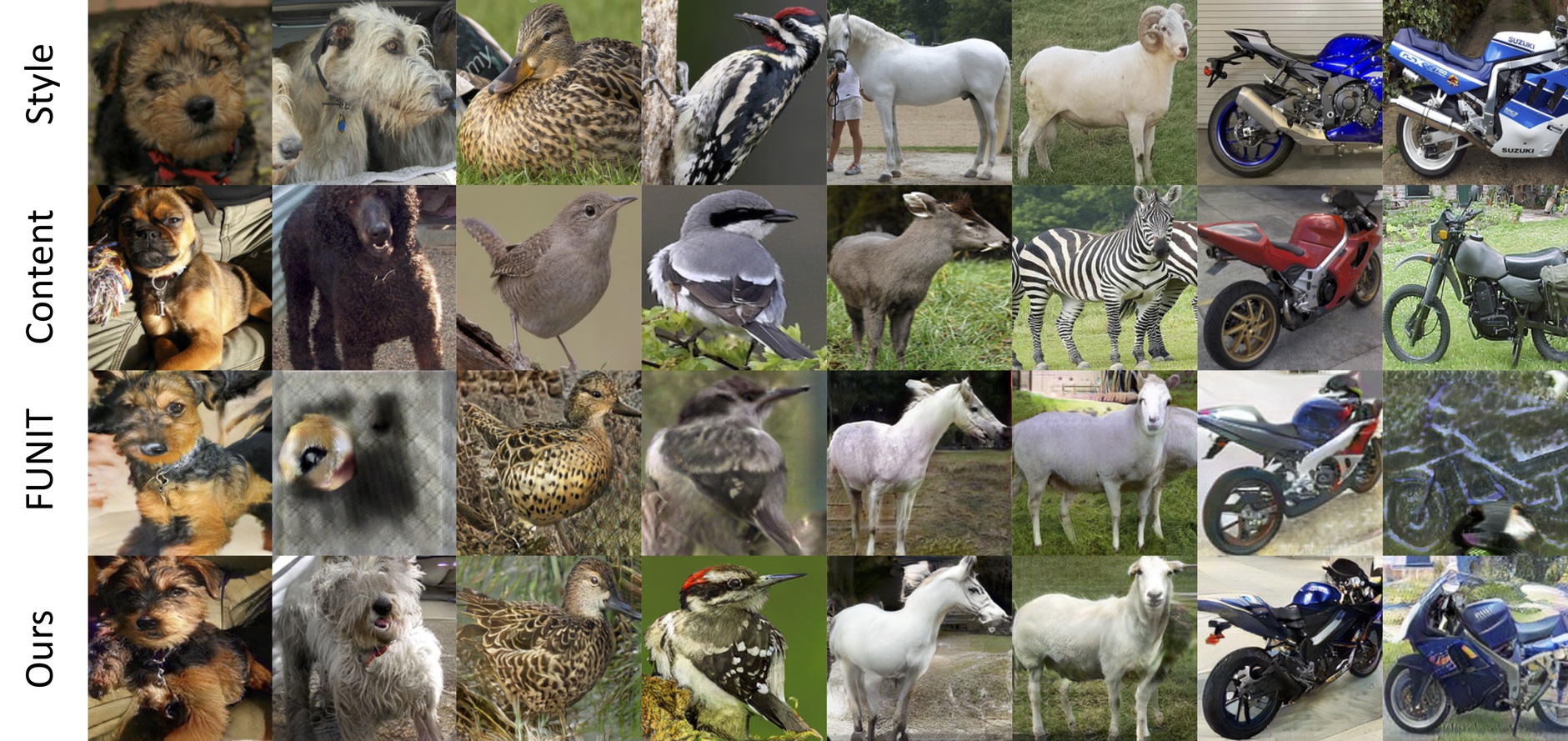}
    \caption{Results on one-shot image-to-image translation. Column 1 \& 2 are from the Carnivores dataset. Column 3 \& 4 are from the Birds dataset. Column 5 \& 6 are from the Mammals dataset. Column 7 \& 8 are from the Motorbikes dataset.}
    \label{fig:one_shot}
\end{figure}

\mysec{Experiments}
We evaluate our method on several challenging datasets that contain large pose variations, part variations, and category variations. Unlike the \funit work, which focuses on translations between reasonably-aligned images or simple objects, our interest is in the translations between likely misaligned images of highly articulate objects. Throughout the experiments, we use 256$\times$256 as our default image resolution for both inputs and outputs.

\mysubsec{Implementation.} We use Adam~\cite{kingma2014adam} with $lr=0.0001$, $\beta_{1}=0.0$, and $\beta_{2}=0.999$ for all methods. Spectral normalization~\cite{miyato2018spectral} is applied to the discriminator. The final generator is a historical average version of the intermediate generators~\cite{karras2017progressive} where the update weight is 0.001. We train the model for 150,000 iterations in total. For every competing model, we compute the scores every 10,000 iterations and report the scores of the iteration that achieves the smallest mFID. Each training batch consists of 64 content images, which are evenly distributed on a DGX machine with 8 V100 GPUs, each with 32GB RAM.

\begin{table}[t]
\begin{center}
\caption{Results on the benchmark datasets.}
\label{tb:vsfunit}
\begin{tabular}{c|c||ccc||c|c}
\toprule[.5pt]
\multirow{2}{*}{Dataset}& \multirow{2}{*}{Method} & \multirow{2}{*}{mFID $\downarrow$}&\multirow{2}{*}{PAcc $\uparrow$}&\multirow{2}{*}{mIoU $\uparrow$}& User Style & User Content\\
&  &  && & Preference $\uparrow$& Preference $\uparrow$\\\hline

\multirow{2}{*}{Carnivores}&FUNIT &147.8&59.8&44.6& 16.5&11.9\\ 
&Ours&\bf{107.8}&\bf{66.5}&\bf{52.1}&\bf{83.5} & \bf{88.1}\\\hline

\multirow{2}{*}{Mammals}&FUNIT &245.8&35.3&23.3 &23.6&27.8\\
&Ours &\bf{109.3}&\bf{48.8}&\bf{35.5}&\bf{76.4}&\bf{72.2}\\\hline

\multirow{2}{*}{Birds}&FUNIT &89.2&52.4&37.2&38.5&37.5\\
&Ours&\bf{74.6}&\bf{53.3}&\bf{38.3}&\bf{61.5}&\bf{62.5}\\\hline

\multirow{2}{*}{Motorbikes}&FUNIT &275.0&85.6&73.8&17.8&17,4\\ 
&Ours &\bf{56.2}&\bf{94.6}&\bf{90.3}&\bf{82.2}&\bf{82.6}\\
  \bottomrule[.5pt]
\end{tabular}
\end{center}

\end{table}
\begin{table}[!t]
	\begin{center}
		\caption{Ablation study on the Carnivores and Birds dataset. "Ours w/o CC" represents a baseline where the content conditioning part in COCO is removed. "Ours w/o CSB" represents a baseline where the CSB is removed. Detailed architecture of these baselines are given in Appendix~\ref{sec::arch}}
		\label{tb:ablation}
		\begin{tabular}{c|c|c|c|c|c|c}
			\toprule[0.5pt]
			\multirow{2}{*}{Method}& \multicolumn{3}{c|}{Carnivores} &  \multicolumn{3}{c}{Birds} \\
			 &mFID$\downarrow$&PAcc$\uparrow$&mIou $\uparrow$ &mFID$\downarrow$&PAcc$\uparrow$&mIou $\uparrow$\\\hline
			Ours w/o \coco &\bf{99.6}&62.5&47.8 &\bf{68.8}&52.8&37.9 \\
			Ours w/o CSB &  107.1&61.8&46.9 &74.1&52.5&37.7\\
			Ours w/o CC &110.0&\bf{66.7}&\bf{52.1} &75.3&52.8&37.9\\
			Ours & 107.8&66.5&\bf{52.1} &74.6&\bf{53.3}&\bf{38.3} \\
			\bottomrule[0.5pt]
		\end{tabular}
	\end{center}
\end{table}

\mysubsec{Datasets.} We benchmark our method using 4 datasets. Each of the dataset contains objects with diverse poses, parts, and appearances.
\begin{itemize}[label=\textbullet, topsep=2pt, itemsep=2pt, leftmargin=*]
	
\item \textit{Carnivores.} We build the dataset using images from the ImageNet dataset\cite{imagenet}. We pick up images from the 149 carnivorous animals and used 119 as the source/seen classes and 30 as the target/unseen classes.

\item \textit{Mammals.} We collect 152 classes of herbivore animal images using Google image search and combine them with the Carnivores dataset to build the Mammals dataset. Out of the 301 classes, 236 classes are used for the source/seen and the rest is used for the target/unseen.

\item \textit{Birds.} We collect 205 classes of bird images using Google image search. 172 classes are used for training and the rest is used for the testing.
 
\item \textit{Motorbikes.} We also collected 109 classes of motorbike images in the same way. 92 classes are used as the source and the rest is used for the target.
\end{itemize}

\begin{figure}[!tbh]
    \centering
    \includegraphics[width=.99\textwidth]{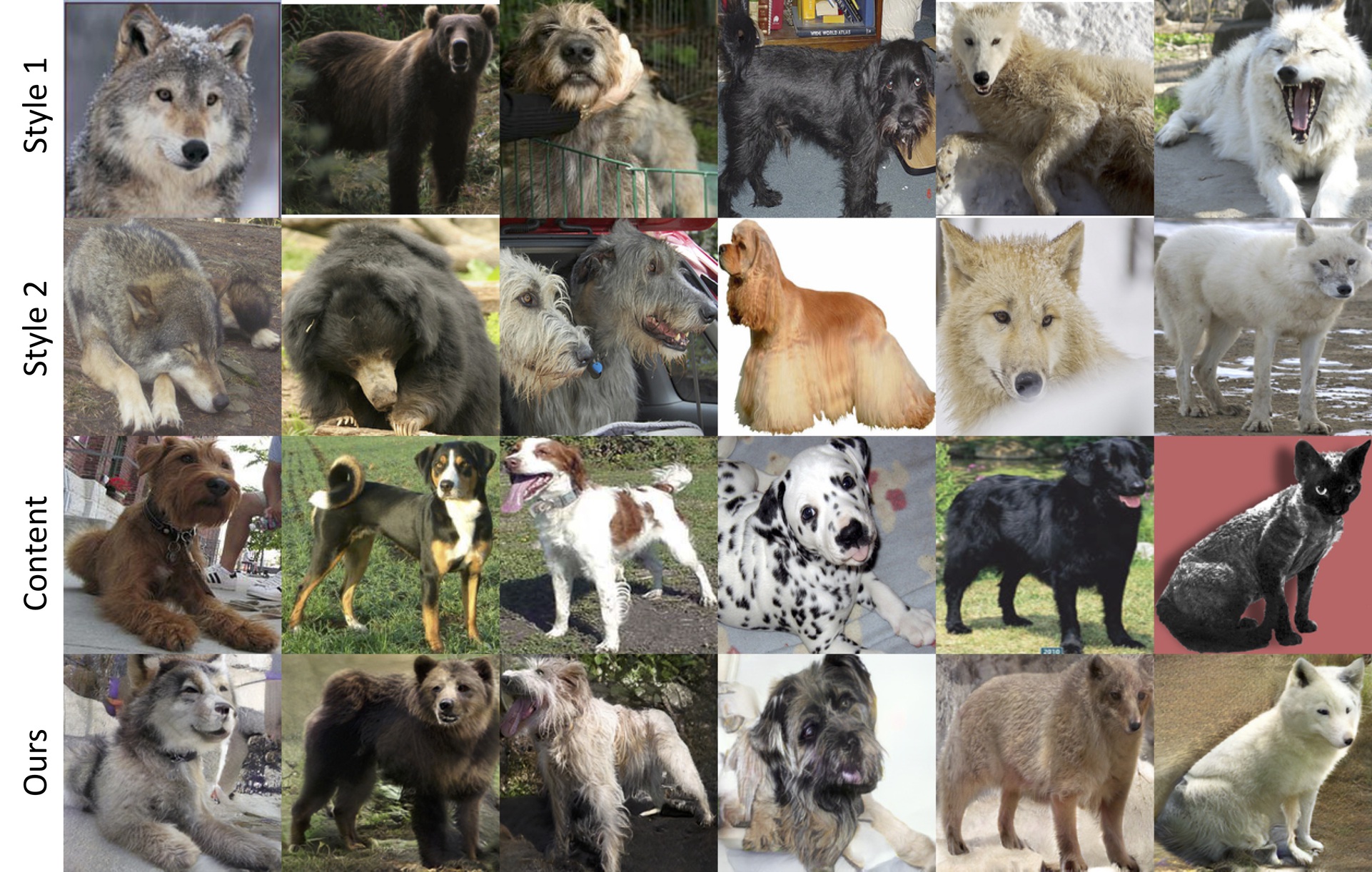}
    \caption{Two-shot image translation results on the Carnivores dataset.}
    \label{fig:two_shot_carnivoroes}
\end{figure}

\begin{figure}[!tbh]
    \centering
    \includegraphics[width=.99\textwidth]{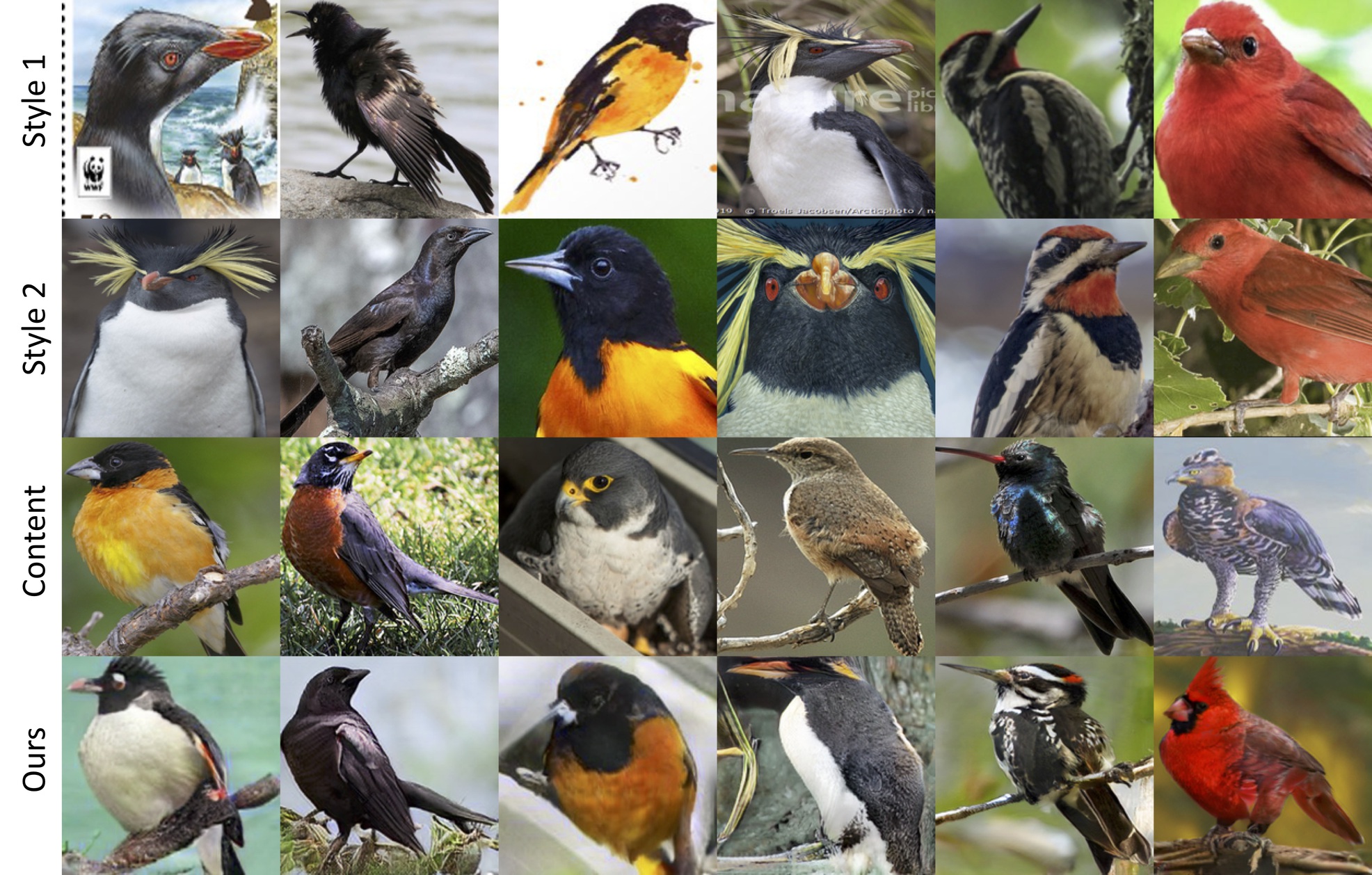}
    \caption{Two-shot image translation results on the Birds dataset.}
    \label{fig:two_shot_bird}
\end{figure}

\begin{figure}[!tbh]
    \centering
    \includegraphics[width=.99\textwidth]{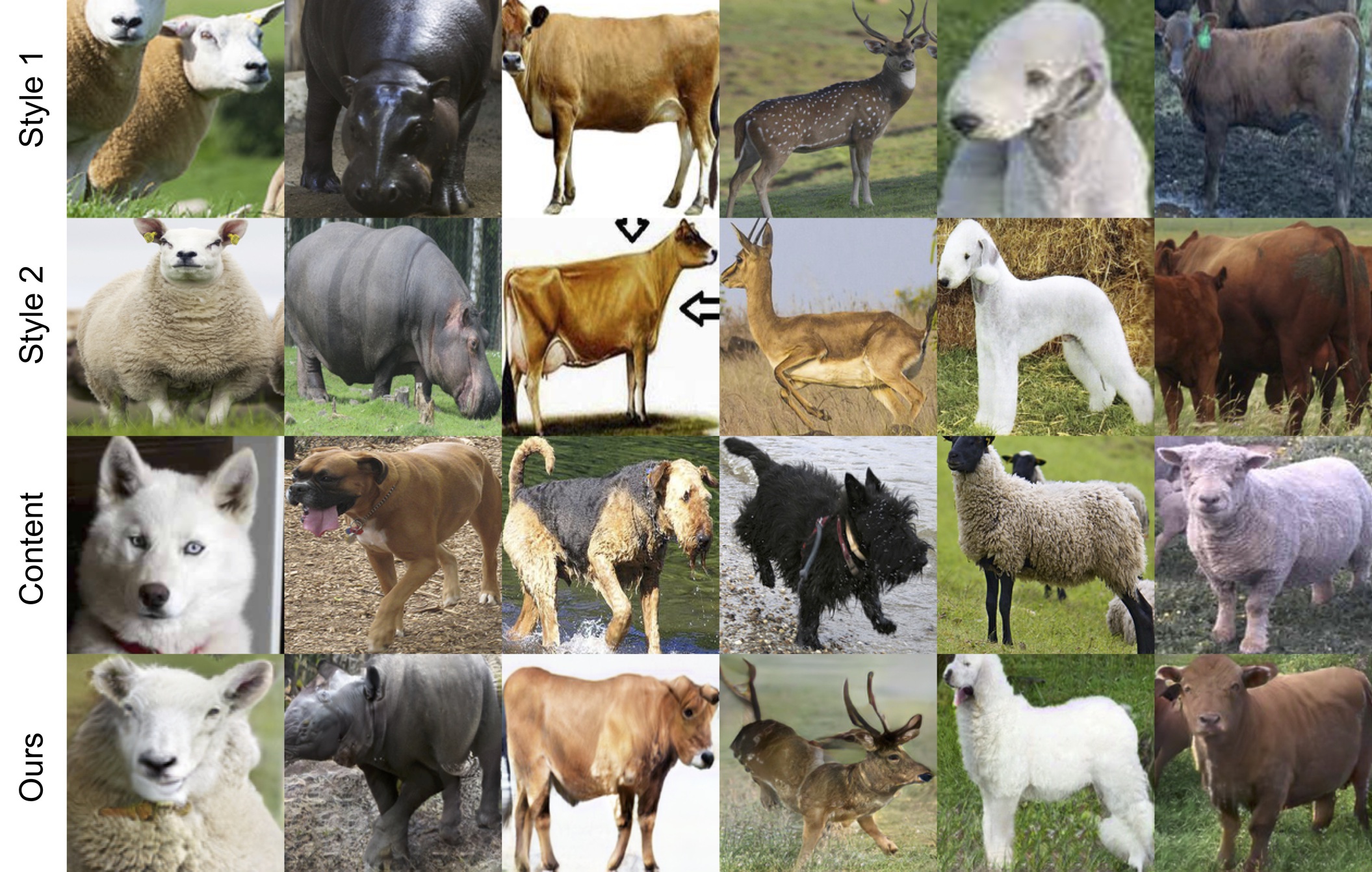}
    \caption{Two-shot image translation results on the Mammals dataset.}
        \label{fig:two_shot_mammal}
\end{figure}

\mysubsec{Evaluation protocol.} For each dataset, we train a model using the source classes mentioned above and test the performance on the target classes for each competing methods. In the test phase, we randomly sample 25,000 content images and pair each of them with a few style images from a target class to compute the translation. Unless specified otherwise, we use the one-shot setting for performance evaluation as it is the most challenging few-shot setting. We evaluate the quality of the translated images using various metrics as explained below.

\mysubsec{Performance metrics.} Ideally, a translated image should keep the structure of the input content image, such as the pose or scale of body parts, unchanged when emulating the appearances of the unseen domain. Existing work mainly focused on the style transfer evaluation because the experiments are performed on well-aligned images or images of simple objects. To consider both the style translation and content preservation, we employ the following metrics. First, we evaluate the style transfer by measuring distance between the distribution of the translated images and the distribution of the real images in the unseen domain using mFID. Second, the content preservation is evaluated by measuring correspondence between a content and a translated image by matching their segmentation masks using mIou and PAcc. Third, we conduct a user study to compute human preference scores on both the style transfer and content preservation of the translation results. The details of the performance metrics are given in Appendix~\ref{sec::metrics}.

\mysubsec{Baseline.} We compare our method with the \funit method because it outperforms many baselines with a large margin as described in Liu~\etal~\cite{liu2019few}. Therefore, a direct comparison with this baseline can verify the effectiveness of the proposed method for the few-shot image-to-image translation task. 

\mysubsec{Main results.} The comparison results is summarized in Table~\ref{tb:vsfunit}. As shown, our method outperforms \funit by a large margin in all the datasets on both automatic metrics and human preference scores. This validates the effectiveness of our method for few-shot unsupervised image-to-image translation. Fig.~\ref{fig:content_loss} and~\ref{fig:one_shot} compare the one-shot translation results computed by the \funit method and our approach. We find images generated by the \funit method contain many artifacts while our method can generate photorealistic and faithful translation outputs. In Fig.~\ref{fig:two_shot_carnivoroes},~\ref{fig:two_shot_bird}, and~\ref{fig:two_shot_mammal}, we further visualize two-shot translation results. More visualization results are provided in Appendix~\ref{sec::additional_results}.
 
 \mysubsec{Ablation study}. In Table~\ref{tb:ablation}, we ablate modules in our architecture and measure their impact on the few-shot translation performance using the Carnivores and Birds datasets. Now, let us walk through the results. First, we find using the CSB improve content preservation scores ("Ours w/o CSB" vs "Ours"), reflected by the better PAcc and mIoU scores achieved. Second, using content conditioning improves style transferring ("Ours w/o CC" vs "Ours"), reflected by the better mFID scores achieved. We also note that despite "Ours w/o COCO" achieves a better mFID, it is in the expense of large content loss. We note that the numbers of parameters of the translation model are 241M for Ours w/o \coco and 242M for Ours. Therefore, adding the \coco module only increases the size of the network by a tiny amount.
 
 In Appendix~\ref{sec::additional_results}, we present additional ablation study on the network architecture design choices. We also analyze impacts of the translation performance with respect to the number of example images available at test time and the number of style classes available at the training time.

\begin{figure}[!t]
    \centering
    \includegraphics[width=\textwidth]{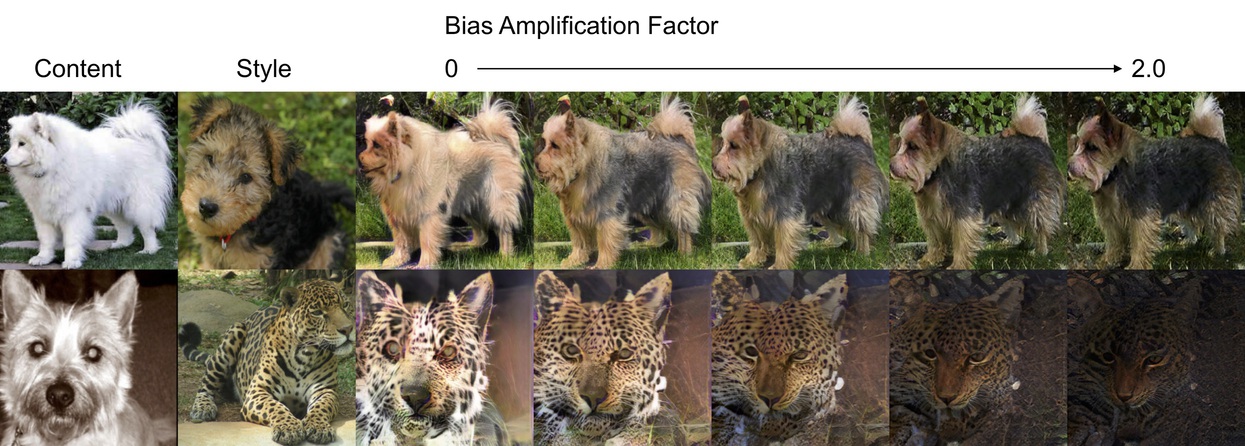}
    \caption{By changing the amplification factor $\lambda$ of the CSB, our model generates different translation outputs for the same pair of content and style images.}
    \label{fig:csb_manipulation}
\end{figure}

\mysubsec{Effect of the CSB}. We conduct an experiment to understand how the CSB designed added to our COCO influences the translation results. Specifically, during testing, we multiply the CSB with a scalar $\lambda$. We then change the $\lambda$ value to visualize its effect as shown in Fig.~\ref{fig:csb_manipulation}. Interestingly, different values of $\lambda$ generate different translation results. When the value is small, the model mostly changes the texture of the content image. With a large $\lambda$ value, both the shape and texture are changed.

\begin{figure}[!t]
    \centering
    \includegraphics[width=\textwidth]{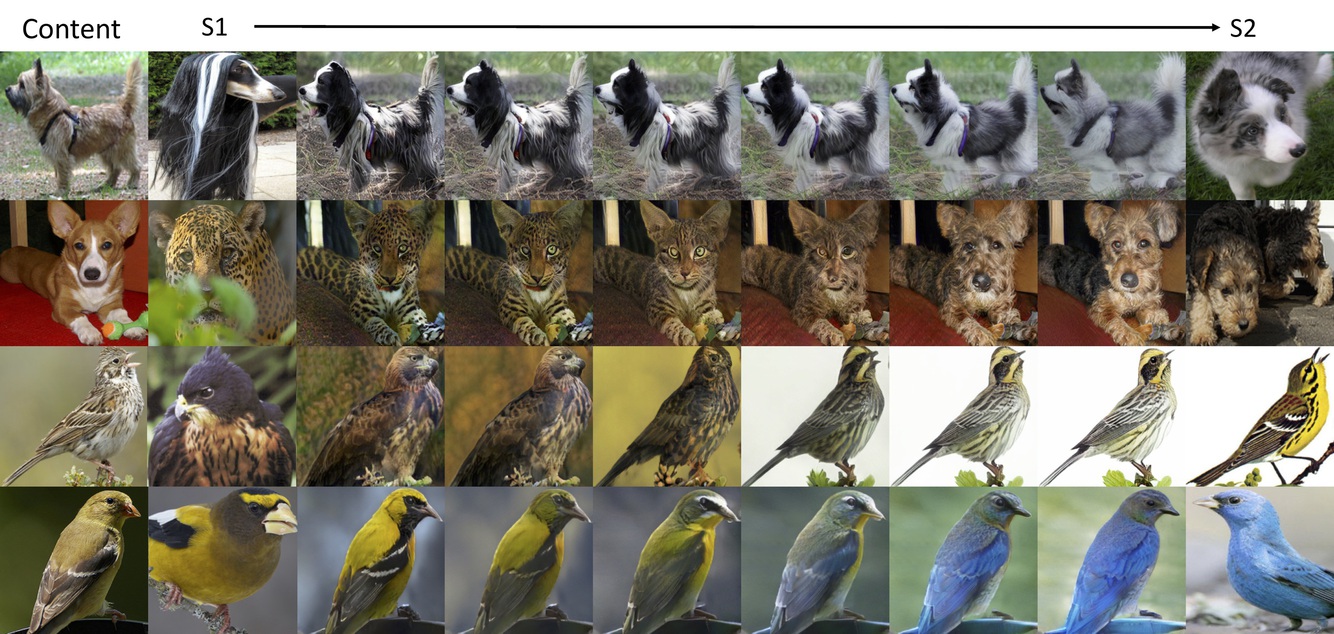}
    \caption{We interpolate the style codes from two example images from two different unseen domains. Our model can generate photorealistic results using these interpolated style codes. More results are in the supplementary materials.}
    \label{fig:blend_carv}
\end{figure}

\mysubsec{Unseen style blending}. Here, we show an application where we combine two style images from two unseen domains to create a new unseen domain. Specifically, we first extract two style codes from two images from two different unseen domains. We then mix their styles by linear interpolating the style codes. The results are shown in Fig. \ref{fig:blend_carv} where the leftmost image is the content and row indicated by \texttt{S1} and \texttt{S2} are the two style images. We find the intermediate style codes render plausible translation results.

\mysubsec{Failure cases}. While our approach effectively addresses the content loss problem, it still have several failure modes. We discuss these failure modes in Appendix~\ref{sec::additional_results}.

\mysec{Conclusion}
We introduced the \cocofunit architecture, a new style encoder for few-shot image-to-image translation that extracts the style code from the example images from the unseen domain conditioning on the input content image and uses a constant style bias design. We showed that the \cocofunit can effectively address the content loss problem, proven challenging for few-shot image-to-image-translation.

\mysubsec{Acknowledgements.} We would like to thank Jan Kautz for his insightful feedback on our project. Kuniaki Saito and Kate Saenko were supported by Honda, DARPA and NSF Award No. 1535797. Kuniaki Saito contributed to this work during his internship at NVIDIA.
\bibliographystyle{splncs}
\bibliography{gan}
\vfill

\appendix
\mysec{Network Architecture}\label{sec::arch}

Here, we present the design of our various sub-networks described in the main paper. Note that we apply the ReLU nonlinearity to all the convolutional layers in the sub-networks in the generator.

\begin{figure*}[!tbh]
    \centering
    \includegraphics[width=0.7\linewidth]{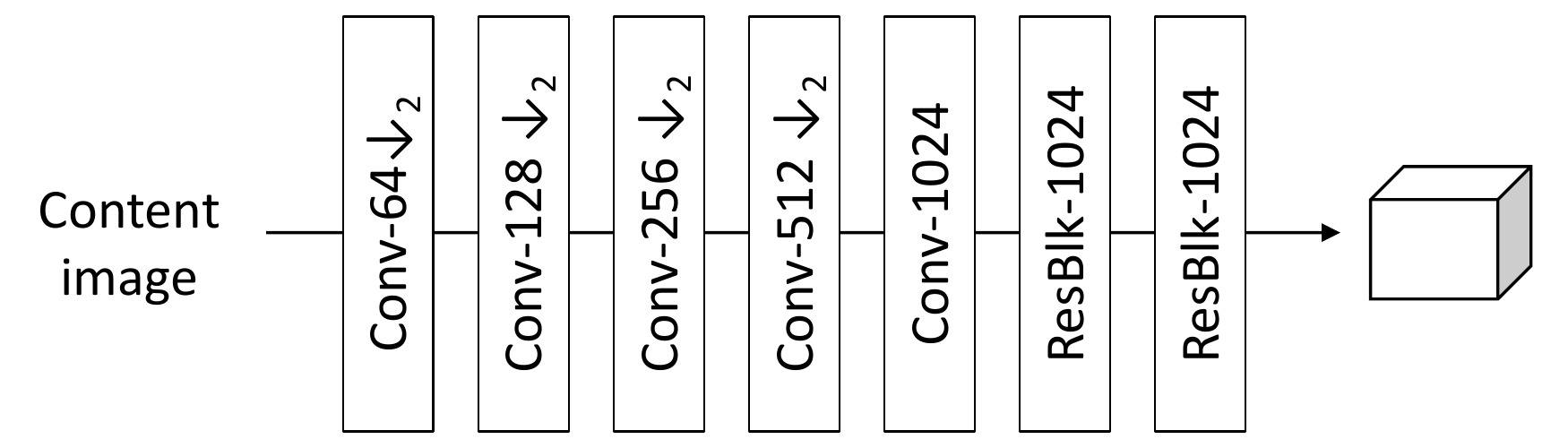}
    \caption{Architecture of the content encoder $E_c$.}
    \label{fig:content_encoder}
\end{figure*}

\mysubsec{Content Encoder.} For the content encoder $E_c$, which is used to obtain a content code to be fed into the decoder for image translation, we mostly follow the content encoder architecture in \funit~\cite{liu2019few}. The only modification we apply is to add one more downsampling layer since the image resolution used in the experiments in the \funit paper is 128$\times$128 and the image resolution in our experiments is 256$\times$256. The detailed design is given in Fig.~\ref{fig:content_encoder}.

\begin{figure*}[!tbh]
    \centering    
    \includegraphics[width=0.7\linewidth]{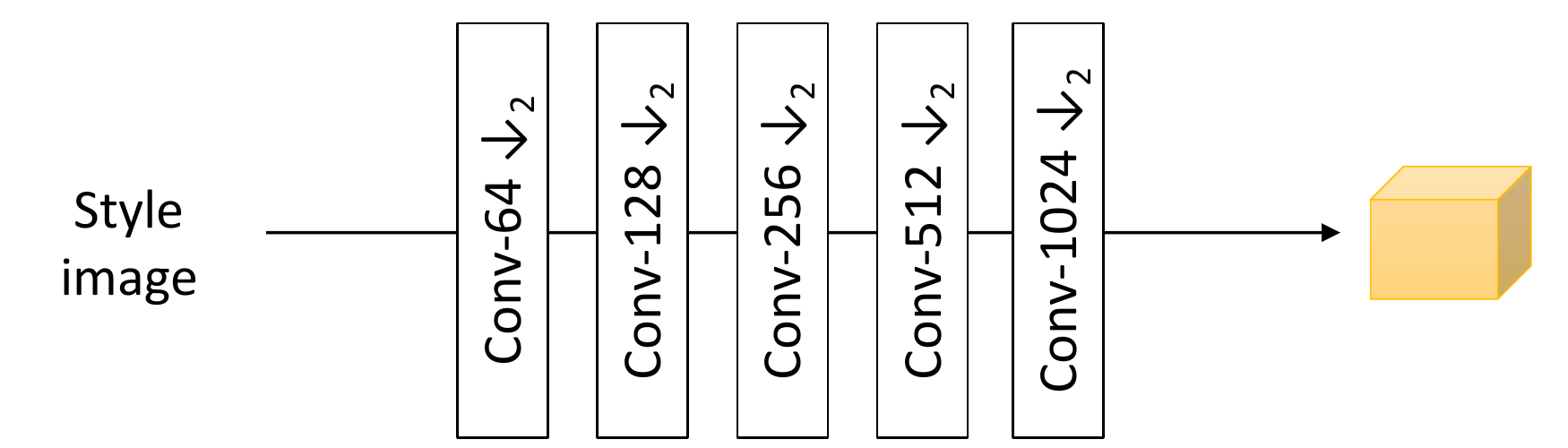}
	\caption{Architecture of the sub-network $E_{s,s}$ in the content-conditioned style encoder (COCO).}
	\label{fig:style_encoder}
\end{figure*}

\mysubsec{Content-Conditioned Style Encoder (COCO).} In COCO, we have two sub-networks, $E_{s,c}$ and $E_{s,s}$ as shown in Fig.~4 of the main paper. Similar to $E_c$, $E_{s,c}$ is used to extract content information from the input image. The difference is that the content code extracted by $E_{s,c}$ is used to compute the style code. In our design, to reduce the number of parameters, we let $E_{s,c}$ be identical to $E_c$. For $E_{s,s}$, we follow the design in \funit~\cite{liu2019few} as shown in Fig.~\ref{fig:style_encoder}.

\begin{figure*}[!tbh]
    \centering    	
    \includegraphics[width=0.7\linewidth]{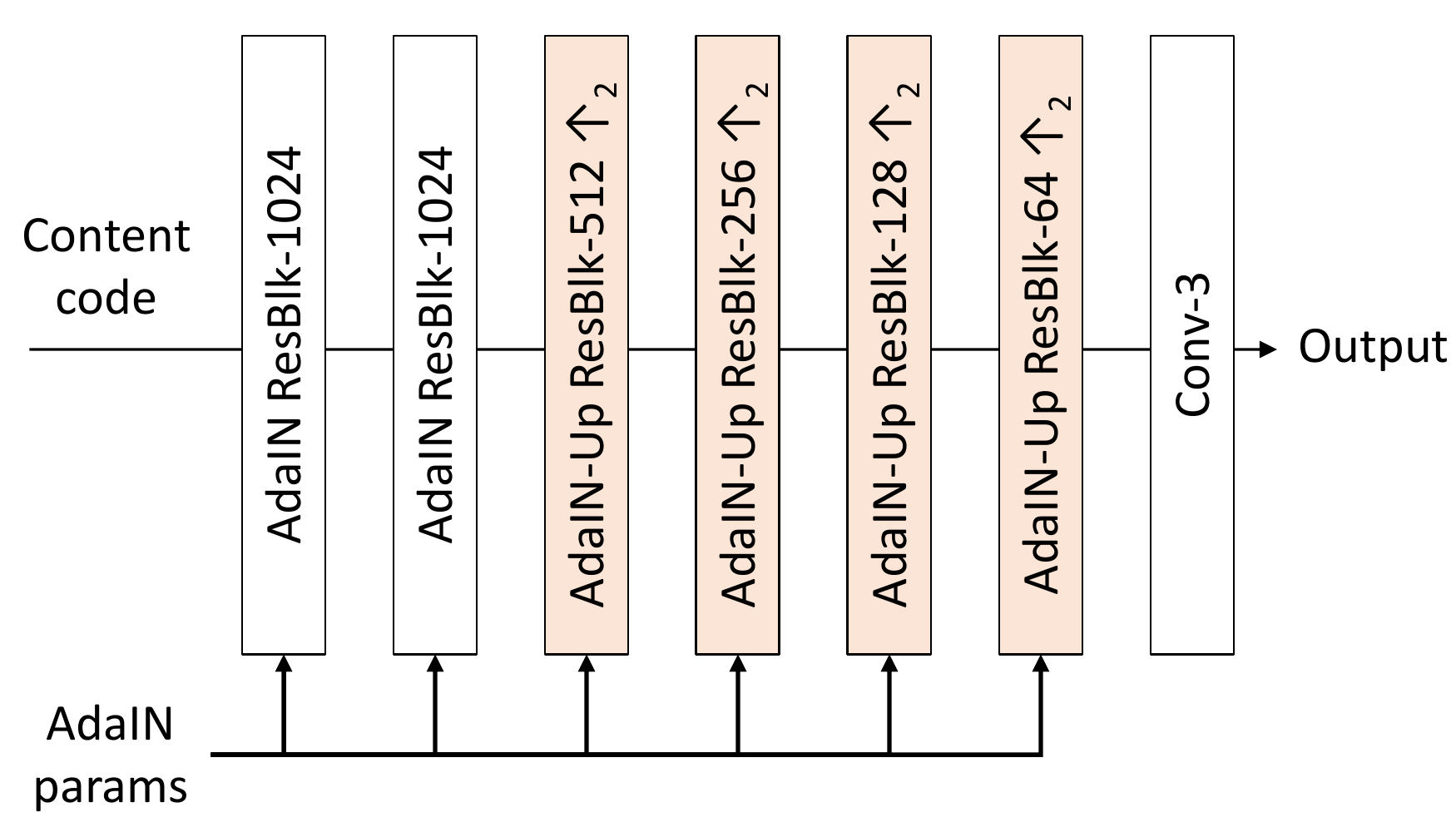}
	\caption{Architecture of the image decoder $F$.}
	\label{fig:decoder}
\end{figure*}    
\begin{figure*}[!tbh]
    \centering    	
    \includegraphics[width=0.7\linewidth]{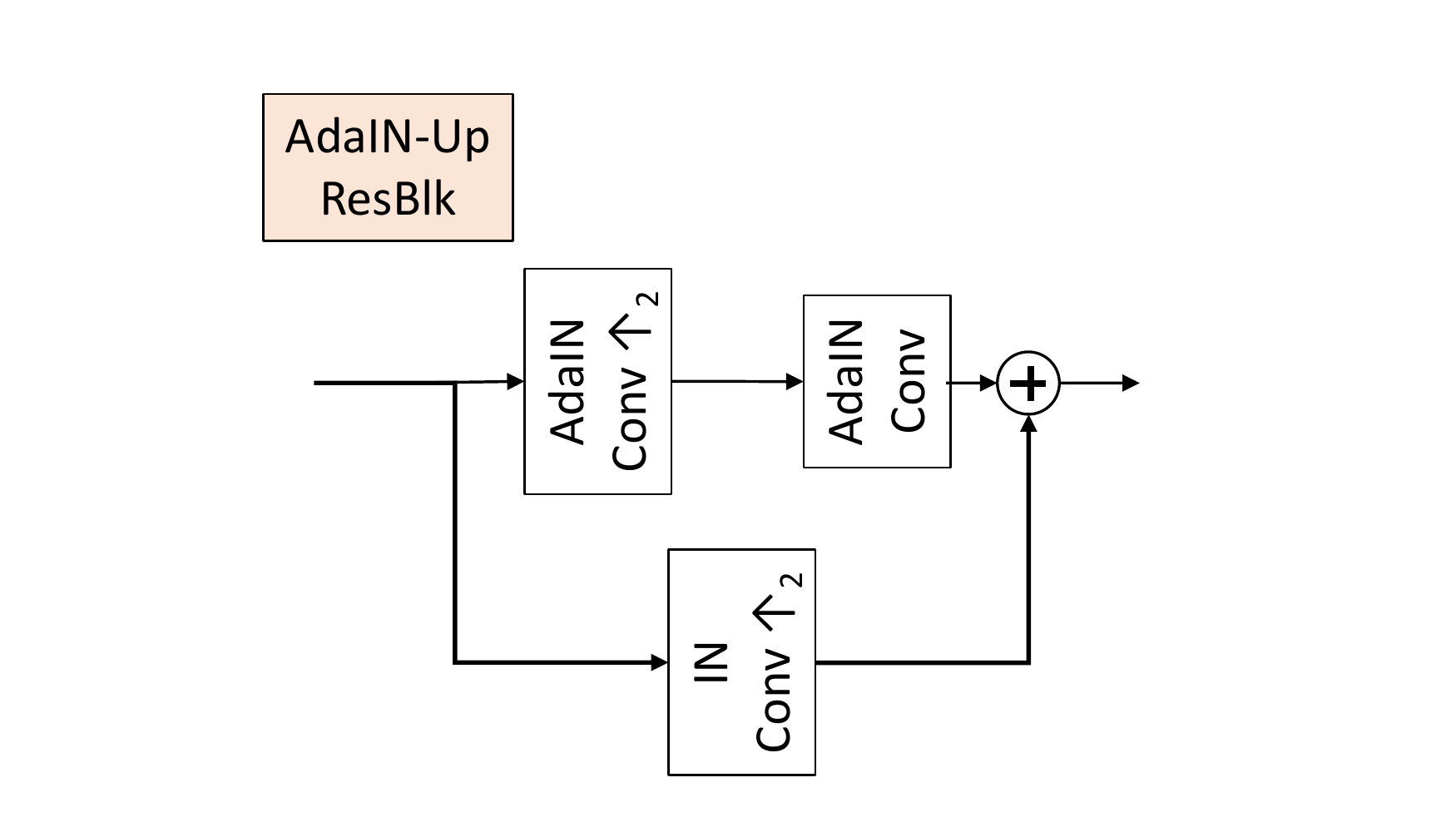}
	\caption{Architecture of the AdaIN-Up residual block.}
	\label{fig:adainup}    
\end{figure*}

\mysubsec{Image decoder.} We propose several modifications to the image generator in \funit. Specifically, we replace several convolutional blocks using a new kind of residual block, which we will call the AdaIN-Up Residual Block. We visualize the image decoder in Fig.~\ref{fig:decoder} where the modification is highlighted in the light orange color. The detail of the AdaIN-Up Residual Block is visualized in Fig.~\ref{fig:adainup}. The residual block consists of a skip connection with upsampling convolution layer followed by an instance normalization layer~\cite{ulyanov2016instance} and a upsampling convolutions with AdaIN~\cite{huang2017adain}.

\mysubsec{Discriminator.} Our discriminator is a patch-based projection discriminator~\cite{miyato2018cgans}. It utilizes the Leaky ReLU nonlinearity. The spectral normalization~\cite{miyato2018spectral} is utilized in every layer. The discriminator consists of one convolutional layer followed by 10 activation first residual blocks~\cite{he2016deep}. The last layer is modified for class conditional projection projection. 
The architecture is illustrated via the following chain of
operations:\\
\texttt{Conv-64 $\rightarrow$ ResBlk-128 $\rightarrow$ ResBlk-128 $\rightarrow$ AvePool2x2 $\rightarrow$ ResBlk-256 \\
	$\rightarrow$ ResBlk-256 $\rightarrow$ AvePool2x2 $\rightarrow$ ResBlk-512 $\rightarrow$ ResBlk-512 \\ 
	$\rightarrow$ AvePool2x2$\rightarrow$ ResBlk-1024 $\rightarrow$ ResBlk-1024 $\rightarrow$ AvePool2x2 \\ 
	$\rightarrow$ ResBlk-1024 $\rightarrow$ ResBlk-1024 $\rightarrow$ (Conv-$1$, Conv-$||\mathbb{S}||$)} where $||\mathbb{S}||$ is the number of source classes.
The last \texttt{(Conv-$1$, Conv-$||\mathbb{S}||)$} denotes the project operation of the patch-based hidden representation and the class embedding.

\section{Learning Objective Function}\label{sec::learning}

We train our model using a similar objective function as in the \funit work. Below, we first describe the individual objective terms and then present the overall optimization problem. Note that we do not utilize the gradient penalty term~\cite{mescheder2018training} used in the \funit work.

\mysubsec{Adversarial loss}.
$\mathcal{L}_{\text{GAN}}(D,G)$ denotes class conditional GAN loss. We use the loss to ensure both photorealism and domain-faithfulness of image translation. Following the projection discriminator design~\cite{miyato2018cgans}, we use the hinge version of the adversarial loss. 

\mysubsec{Reconstruction loss}. 
The loss encourages the model to reconstruct images when both the content and the style are from the same domain. This loss helps regularize the learning and is given by 
\begin{equation}
\mathcal{L}_{\text{R}} (G)= E_{x_c}[||x_c - \bar{\bm{x_c}}||_1], 
\end{equation}
where $\bar{\bm{x_c}} = G (x_c, x_c)$. 

\mysubsec{Discriminator Feature matching loss}.
Minimizing the feature distance between real and fake samples in the discriminator feature space can stabilize the training of adversarial learning and contribute to the performance of a model as used in the \funit work. We take the feature before the last linear layer and performed spatial average pooling. Let the feature computing function as $D_f$. The feature matching loss is given by 
\begin{equation}
\mathcal{L}_{\text{FM}} (G) =  E_{x_s, x_c}[||D_f(x_s) - D_f(\bar{\bm{x_s}})||_1], 
\end{equation}
where $\bar{\bm{x_s}} = G (x_c, x_s)$. 

Overall our training objective is given by, 
\begin{align}
\min_{D}\max_{G} \mathcal{L}_{\text{GAN}}(D,G) + 
\lambda_{\text{R}} \mathcal{L}_{\text{R}}(G) + \lambda_{\text{F}} \mathcal{L}_{\text{FM}}(G), \label{eqn::learning} 
\end{align}
where $\lambda_{\text{R}}$ and $\lambda_{\text{F}}$ denote trade-off parameters for two losses. We set $\lambda_{\text{R}}$ 0.1 and $\lambda_{\text{F}}$ 1.0 in all experiments.

\mysec{Performance Metrics}\label{sec::metrics}

Here, we describe the details of the evaluation metrics that we use to measure style faithfulness, content preservation, and human preference of the translation outputs.

\mysubsec{Style faithfulness.} We use the Frechet Inception Distance (FID)~\cite{heusel2017gans} to measure distance between the distribution of the translated images and the distribution of real unseen images, based on the InceptionV3~\cite{szegedy2016rethinking} network. We compute FID for each of the target class and report their mean (mFID).

\mysubsec{Content preservation.} An ideal translation should keep the structure of input content image unchanged. We measure the content preservation by comparing the body-part segmentation mask of a content image and that of a translated image. Since we do not have ground-truth body-part annotations, we estimate the body-part segmentation masks by using a DeeplabV3~\cite{chen2017rethinking} network trained on the Pascal body part dataset~\cite{everingham2015pascal}. We note similar approaches are used in the other image synthesis prior  works~\cite{wang2018high,park2019semantic,wang2018video,wang2019few}. We obtain the body-part segmentation masks for both content and translated images. Then, we calculate pixel accuracy (PAcc) and mean intersection-over-union (mIoU) by treating the mask of the content image as the ground-truth annotation.

\mysubsec{Human preference.} Using Amazon Mechanical Turk (AMT), we perform a subjective visual test to gauge the quality of few-shot translation results. We conduct two studies: content preservation and style faithfulness. In the first study, each question contains the content image and the translation results from two competing methods, and the AMT worker is asked to choose which translation result better preserves the pose content from the content image. In the second study, each question contains the style image and the translation results from two competing methods, and the AMT worker is asked to choose which translation result rendering an object resembles more to the one in the style image. We generate 1000 questions per dataset per study. Each question is answered by 3 different AMT workers with a high approval rating. We use the preference score for evaluation. These are the counterpart of the automatic evaluation metrics described above.

\mysec{Experiments (Cont.)}\label{sec::additional_results}

\begin{table}[!tbh]
	\begin{center}
		\caption{Ablation study on the Carnivores and Birds dataset. "ProjD" and "ResBlks+" represent the variant where we replace the mutli-class disciminator with the projection discriminator~\cite{miyato2018cgans} and the variant where we use additional residual blocks in the decoder. "Ours w/o COCO" represents a baseline where COCO is removed. }
		\label{tb:ablation_supple}
		\begin{tabular}{c|c|c|c|c|c|c}
			\toprule[0.5pt]
			\multirow{2}{*}{Method}& \multicolumn{3}{c|}{Carnivores} &  \multicolumn{3}{c}{Birds} \\
			 &mFID$\downarrow$&PAcc$\uparrow$&mIou $\uparrow$ &mFID$\downarrow$&PAcc$\uparrow$&mIou $\uparrow$\\\hline
			Ours w/o COCO w ProjD \& ResBlks+ &147.1&59.9&44.7&89.2&52.4&37.2\\
			Ours w/o COCO w ResBlks+ &114.3&65.7&51.3&75.0&51.9&37.3\\
			Ours w/o COCO w ProjD &138.5&54.2&39.4&97.4&52.6&36.9\\
			Ours w/o COCO &\bf{99.6}&62.5&47.8 &\bf{68.8}&52.8&37.9 \\
			Ours& 107.8&\bf{66.5}&\bf{52.1} &74.6&\bf{53.3}&\bf{38.3} \\
			\bottomrule[0.5pt]
		\end{tabular}
	\end{center}
\end{table}

\mysubsec{More ablation study.} Table ~\ref{tb:ablation_supple} shows the ablation of the proposed style encoder, projection discriminator ~\cite{miyato2018spectral} and additional residual blocks in decoder. We find our full model better preserves content information and achieve better performance on the mFID metric. We also find that projection discriminator and additional residual blocks in the generator helps lower the mFID scores.

\begin{figure*}[!t]
\begin{center}
\begin{subfigure}[b]{0.23\textwidth}
  \begin{center}
   \includegraphics[width=\textwidth]{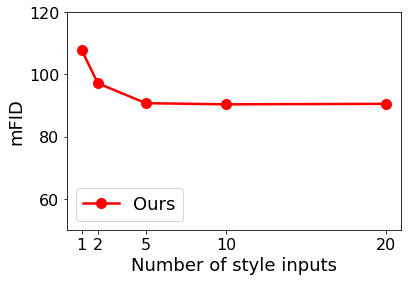}
    \caption{}
     \label{fig:fid_changenum}
  \end{center}
 \end{subfigure}
 ~ 
 \begin{subfigure}[b]{0.23\textwidth}
  \begin{center}
   \includegraphics[width=\textwidth]{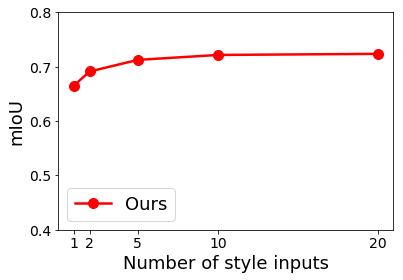}
    \caption{}
    
    \label{fig:iou_changenum}
  \end{center}
 \end{subfigure}
 ~ 
 \begin{subfigure}[b]{0.23\textwidth}
  \begin{center}
   \includegraphics[width=\textwidth]{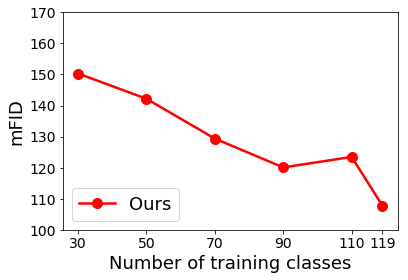}
    \caption{}
     \label{fig:train_class}
  \end{center}
 \end{subfigure}
 ~ 
 \begin{subfigure}[b]{0.23\textwidth}
  \begin{center}
   \includegraphics[width=\textwidth]{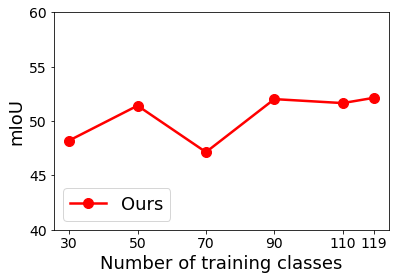}
    \caption{}
     \label{fig:train_class_miou}
  \end{center}
 \end{subfigure}
 ~ 

  \caption{(a)(b): mFID and mIoU scores with respect to varying number of style images respectively. (c)(d): mFID and mIoU with respect to varying number of training classes.}
  \end{center}
 \end{figure*}

\begin{figure*}[!tbh]
\begin{center}
\begin{subfigure}[b]{0.31\textwidth}
  \begin{center}
   \includegraphics[width=\textwidth]{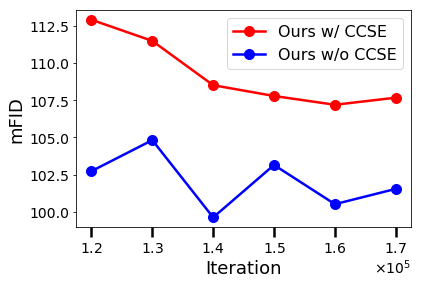}
     \label{fig:carv_fid_iter}
  \end{center}
 \end{subfigure}
 ~ 
 \begin{subfigure}[b]{0.31\textwidth}
  \begin{center}
   \includegraphics[width=\textwidth]{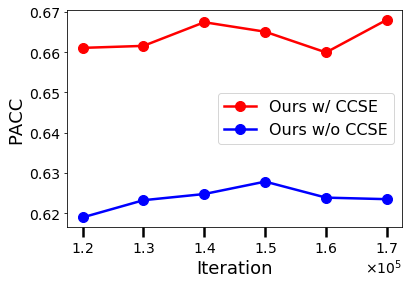}
    \label{fig:carv_pacc_iter}
  \end{center}
 \end{subfigure}
 ~ 
 \begin{subfigure}[b]{0.31\textwidth}
  \begin{center}
   \includegraphics[width=\textwidth]{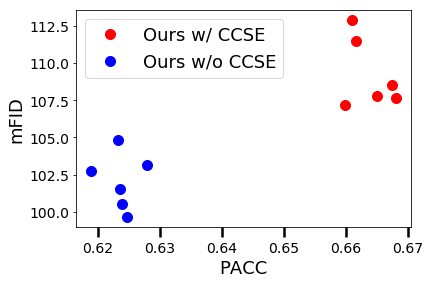}
     \label{fig:carv_mfid_pacc}
  \end{center}
 \end{subfigure}
 ~ 
 \begin{subfigure}[b]{0.31\textwidth}
  \begin{center}
   \includegraphics[width=\textwidth]{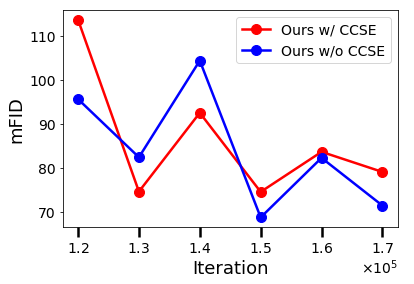}
     \label{fig:bird_fid_iter}
  \end{center}
 \end{subfigure}
 ~ 
 \begin{subfigure}[b]{0.31\textwidth}
  \begin{center}
   \includegraphics[width=\textwidth]{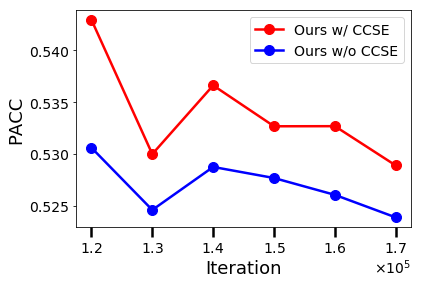}
        \label{fig:bird_pacc_iter}
  \end{center}
 \end{subfigure}
 ~
 \begin{subfigure}[b]{0.3\textwidth}
  \begin{center}
 \includegraphics[width=\textwidth]{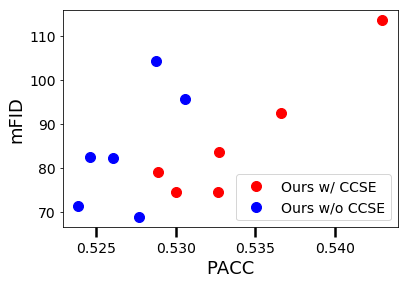}
     \label{fig:bird_mfid_pacc}
  \end{center}
 \end{subfigure}
  \caption{Translation performance vs training iterations. Top row: Carnivorous dataset, bottom row: Bird dataset.}
   \label{fig:curve_carn}
  \end{center}
 \end{figure*}

\mysubsec{Effect on number of examples}. We further investigate the relation between translation performance and number of input style images using the Carnivores dataset. Fig.~\ref{fig:fid_changenum} and \ref{fig:iou_changenum} show that the translation performance of our method are positively correlated with the number of input style images.

\mysubsec{Effect on number of source classes}. Fig. \ref{fig:train_class} and \ref{fig:train_class_miou} show the number of sources classes used in training versus the achieved mFID and mIoU score on the Carnivores dataset. When the model sees more source classes during training, it renders a better few-shot image translation performance during testing. 

\mysubsec{Performance vs training time}. Fig. \ref{fig:curve_carn} shows the training performance vs iterations on the Carnivores and Birds datasets. Both the mFID and PAcc scores improve as training proceeds for the Carnivores dataset. For the Birds dataset, the PAcc score degrades while the mFID improves. For both datasets, ours with COCO have better content preservation score. 

\begin{figure}[!tbh]
    \centering
    \includegraphics[width=\textwidth]{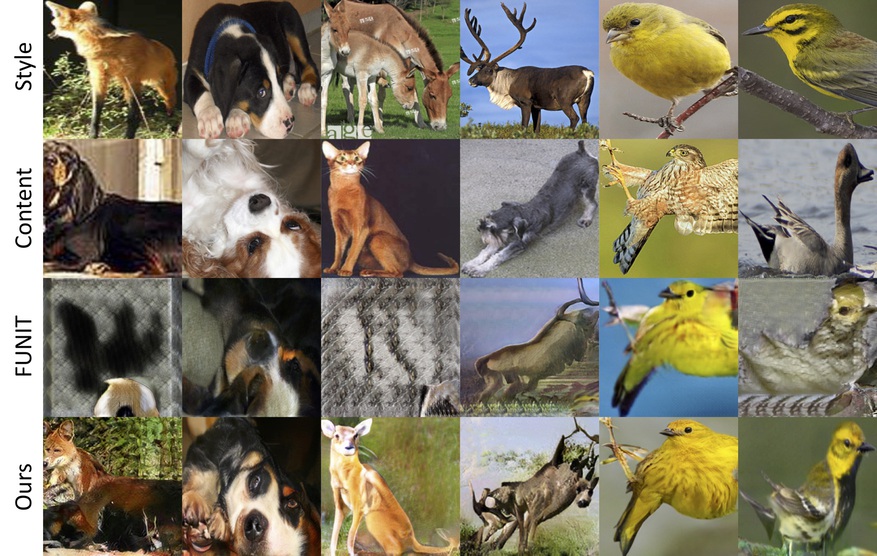}
    \caption{Failure cases. Column 1 \& 2 are from the Carnivores dataset. Column 3 \& 4 are from the Mammals dataset. Column 5 \& 6 are from the Birds dataset.}
    \label{fig:failure}
\end{figure}

\mysubsec{Failure cases}. Fig.~\ref{fig:failure} illustrates several failure cases generated by our method. When the body part of the input content is hard to localize, the model generates incorrect results. Sometimes, the model generates hybrid classes.

\begin{figure}[!tbh]
    \centering
    \includegraphics[width=\textwidth]{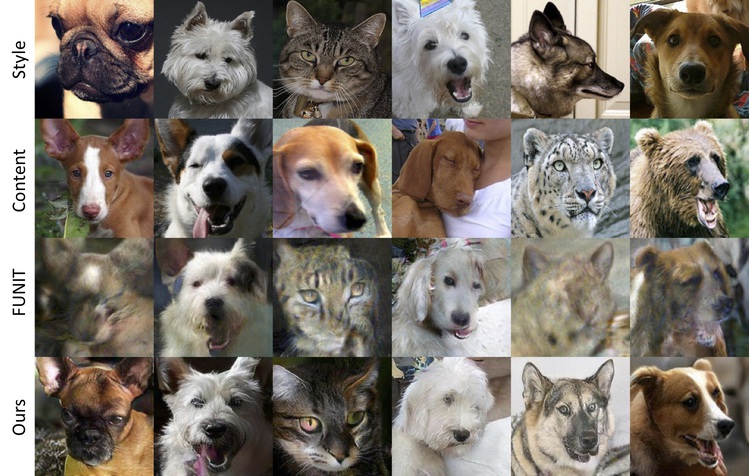}
    \caption{Results on one-shot image-to-image translation for the Animal Faces dataset proposed in \funit~\cite{liu2019few} work.}
    \label{fig:animalface}
\end{figure}

\mysubsec{Results on a well-aligned dataset}. Fig. \ref{fig:animalface} shows the translation results on the Animal Faces dataset presented in FUNIT paper ~\cite{liu2019few}. The images focus on animal face regions, thus they are well-aligned. We used the same training and test split like the original paper but performed translation on 256x256 image resolution. The mFID, PAcc, mIOU of FUNIT are 196.9, 0.505 and 0.344 respectively while ours are {\bf 106.8, 0.617, 0.459}. Even for the well-aligned dataset, we can see the advantage of using our architecture.

\begin{figure}[!tbh]
    \centering
    \includegraphics[width=\textwidth]{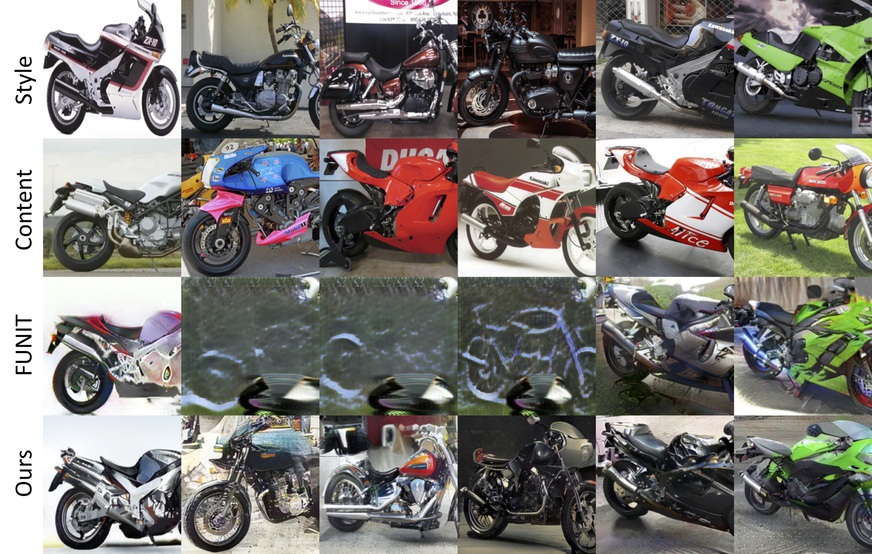}
    \caption{One-shot image-to-image translation results on the Motorbike dataset.}
    \label{fig:bike_supp}
\end{figure}

\begin{figure}[!tbh]
    \centering
    \includegraphics[width=\textwidth]{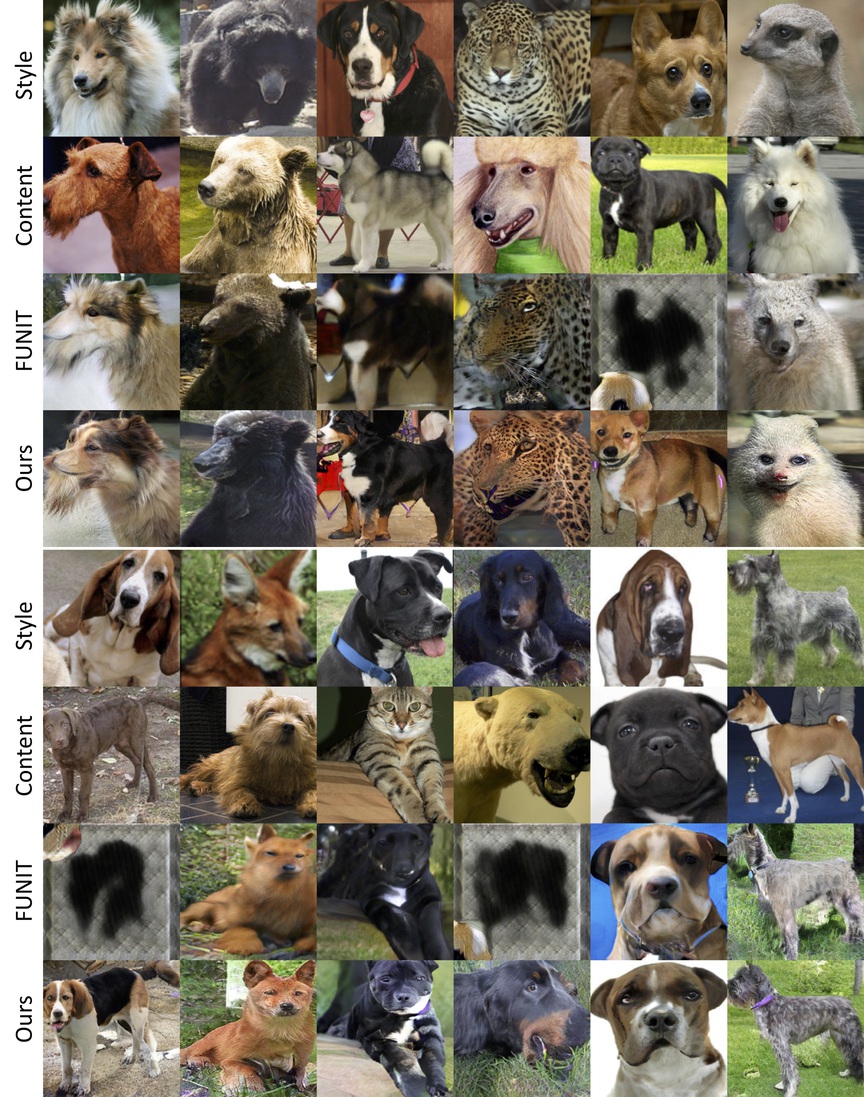}
    \caption{One-shot image-to-image translation results on the Carnivores dataset.}
    \label{fig:carv_supp}
\end{figure}

\begin{figure}[!tbh]
    \centering
    \includegraphics[width=\textwidth]{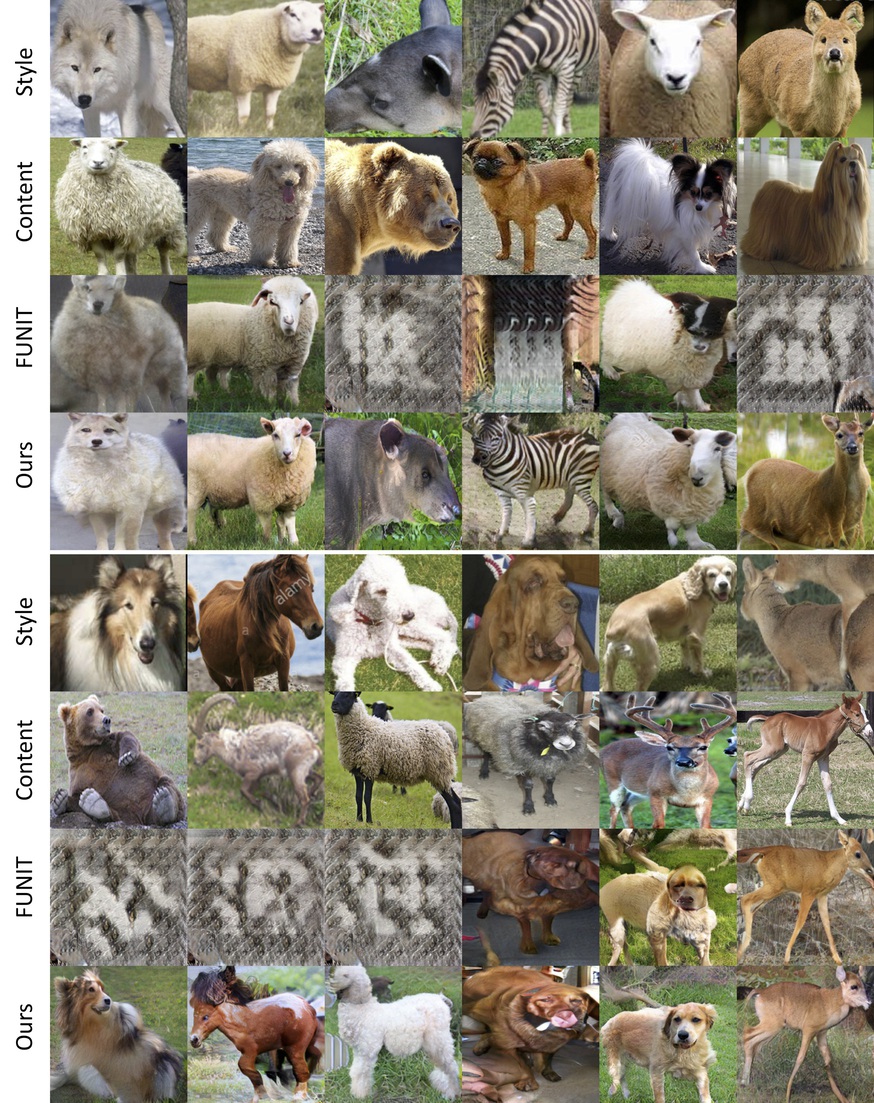}
    \caption{One-shot image-to-image translation results on the Mammals dataset.}
    \label{fig:mammal_supp}
\end{figure}

\begin{figure}[!tbh]
    \centering
    \includegraphics[width=\textwidth]{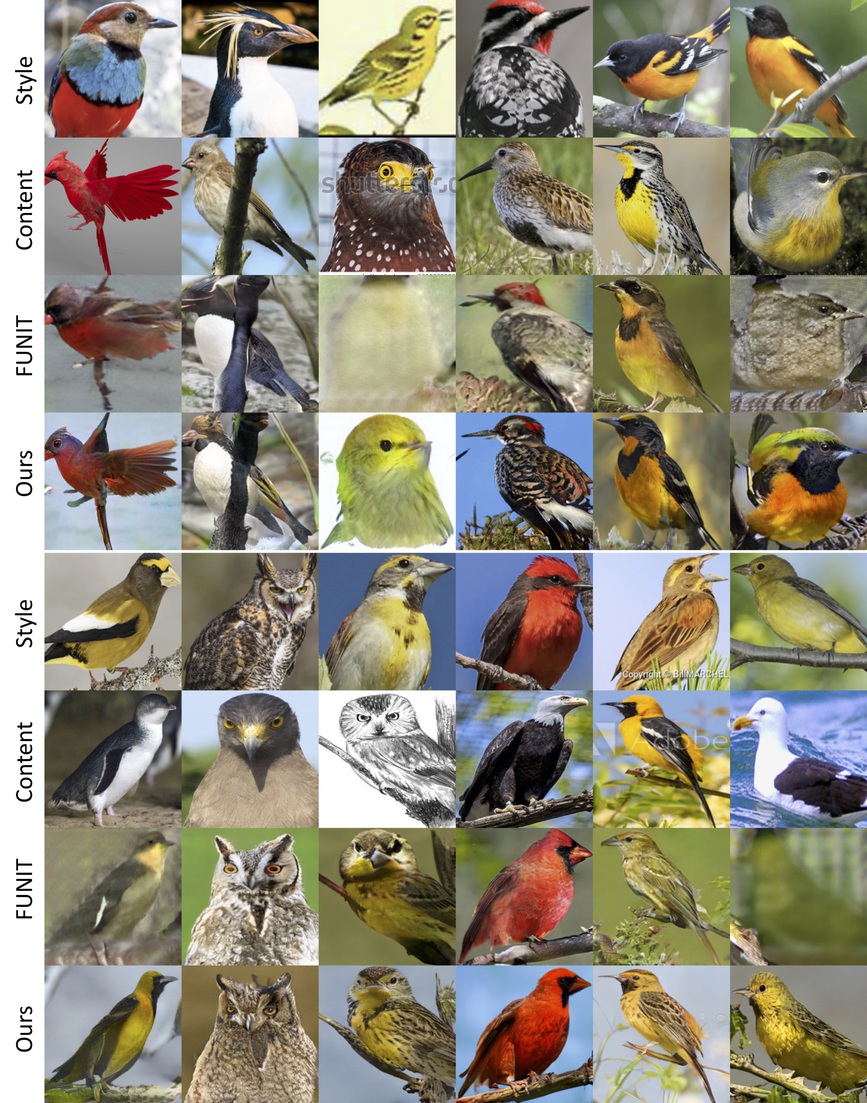}
    \caption{One-shot image-to-image translation results on the Birds dataset.}
    \label{fig:bird_supp}
\end{figure}

\mysubsec{Additional visual results}. Fig. \ref{fig:carv_supp}, \ref{fig:mammal_supp}, \ref{fig:bird_supp}, and \ref{fig:bike_supp} show additional one-shot image translation results on the benchmark datasets. Our model can generate more photorealistic and more faithful translation outputs.

\end{document}